\documentclass{article}
\PassOptionsToPackage{numbers,compress}{natbib}

\PassOptionsToPackage{hidelinks}{hyperref}

\usepackage[preprint]{neurips_2025}

\usepackage{hyperref}
\usepackage{placeins}
\usepackage{amsmath}        
\usepackage{multirow}       
\usepackage{graphicx} 
\usepackage{longtable}

\usepackage[utf8]{inputenc} 
\usepackage[T1]{fontenc}    
\usepackage{url}            
\usepackage{booktabs}       
\usepackage{amsfonts}       
\usepackage{nicefrac}       
\usepackage{microtype}      
\usepackage{xcolor}         

\author{
  Hamed Najafi\\
  Florida International University \\
  Miami, FL \\
  \texttt{hnaja002@fiu.edu} \\
  \AND
  Dongsheng Luo\\
  Florida International University \\
  Miami, FL \\
  \texttt{dlou@fiu.edu} \\ 
  \AND
  Jason Liu\\
  Florida International University \\
  Miami, FL \\
  \texttt{liux@fiu.edu} \\
}
\title{Feature-Function Curvature Analysis: A Geometric Framework for Explaining Differentiable Models}

\begin{document}

\maketitle

\begin{abstract}
 Explainable AI (XAI) is critical for building trust in complex machine learning models, yet mainstream attribution methods often provide an incomplete, static picture of a model's final state. By collapsing a feature's role into a single score, they are confounded by non-linearity and interactions. To address this, we introduce Feature-Function Curvature Analysis (FFCA), a novel framework that analyzes the geometry of a model's learned function. FFCA produces a 4-dimensional signature for each feature, quantifying its: (1) \textbf{Impact}, (2) \textbf{Volatility}, (3) \textbf{Non-linearity}, and (4) \textbf{Interaction}.  Crucially, we extend this framework into \textbf{Dynamic Archetype Analysis}, which tracks the evolution of these signatures throughout the training process. This temporal view moves beyond explaining what a model learned to revealing \emph{how} it learns. We provide the first direct, empirical evidence of hierarchical learning, showing that models consistently learn simple linear effects before complex interactions. Furthermore, this dynamic analysis provides novel, practical diagnostics for identifying insufficient model capacity and predicting the onset of overfitting. Our comprehensive experiments demonstrate that FFCA, through its static and dynamic components, provides the essential geometric context that transforms model explanation from simple quantification to a nuanced, trustworthy analysis of the entire learning process.
\end{abstract}

\section{Introduction}
\label{sec:introduction}

The proliferation of complex, ``black-box'' machine learning models in high-stakes domains has created an urgent need for reliable and insightful explanations of their behavior~\cite{arrieta2020explainable, adadi2018peeking}. In response, the field of Explainable AI (XAI) has developed powerful feature attribution methods like LIME~\cite{ribeiro2016should} and SHAP~\cite{lundberg2017unified}, which assign a single score to quantify a feature's contribution. While widely used, this approach has a fundamental limitation: a single score cannot disambiguate the \emph{nature} of a feature's contribution, conflating direct influence with importance derived from complex non-linearities or interactions~\cite{kumar2020problems}.

However, even \emph{post-hoc} methods—applied after training to interpret a fixed model—though they offer a more detailed, multi-dimensional view of the trained model, still answer only \emph{what} the model learned. They treat the model as a static artifact, ignoring the rich history of the training process that produced it. This leaves a more profound question unanswered: \emph{how} does a model arrive at its final, complex representation? Does it learn simple patterns first? Can we observe the moment it discovers a key interaction? Can we detect when its learning process becomes unstable before its performance degrades? These questions matter for actionable ML practice—they let us diagnose reliance on spurious cues, anticipate failures under distribution shift, audit fairness and safety, and decide where to intervene (e.g., data curation, regularization, or model editing).


In comparison to decomposition methods (e.g., LRP),
Layer-wise Relevance Propagation (LRP) redistributes a prediction backward through the network to assign ``relevance'' to inputs.
In contrast, we introduce \emph{Feature–Function Curvature Analysis (FFCA)}, a post-hoc, model-dependent probe that examines the \emph{local shape} of the trained model in input space.
We call FFCA \emph{geometric} because it characterizes how the model’s output changes in the immediate neighborhood of a data point—capturing not only which features matter, but whether their effects are stable or volatile, nonlinear or approximately linear, and whether they act alone or only in combination with others.
Accordingly, FFCA reports four \emph{diagnostic signatures}: (i) \emph{impact} (how strongly a feature influences the prediction near the point of interest), (ii) \emph{volatility} (how sensitive that influence is to small input changes), (iii) \emph{curvature} (the extent to which the relationship is nonlinear), and (iv) \emph{interaction} (the degree to which a feature’s effect depends on other features).
Thus, while both LRP and FFCA are post-hoc and model-dependent, LRP yields a decomposition of a prediction into additive parts, whereas FFCA yields shape-based diagnostics about the model’s local behavior.

To address these questions, our FFCA framework analyzes both the trained model’s local behavior and its learning dynamics for any differentiable architecture. FFCA dissects a feature's role by analyzing the local geometry of the model's decision surface. It operates in two modes: 1) \textbf{Static Analysis:} At the end of training, FFCA produces a 4-dimensional signature for each feature—\emph{Impact} (strength of local influence), \emph{Volatility} (stability of that influence to small perturbations), \emph{Non-linearity} (departure from a linear response), and \emph{Interaction} (dependence of a feature’s effect on other features). We group these continuous signatures into an \emph{eight-archetype} taxonomy to turn raw numbers into actionable labels, enabling rapid triage, cross-model comparison, and clear communication to practitioners and auditors (with recommended interventions per archetype). 2) \textbf{Dynamic Archetype Analysis:} By capturing the 4D signature at regular intervals during training, FFCA provides a temporal view of the learning process. This allows us to observe the evolution of feature roles, providing unprecedented insight into the model's behavior.


FFCA summarizes local model behavior with four complementary diagnostics that capture strength of influence, stability, non-linearity, and cross-feature dependence (formal definitions in Section~\ref{sec:ffca_methodology}).

The contributions of this paper are threefold:

\begin{enumerate}
    \item We propose FFCA, a novel XAI framework that uses a 4D geometric signature and an 8-archetype taxonomy to provide a nuanced characterization of feature roles.
    \item We introduce Dynamic Archetype Analysis, a major extension that tracks the evolution of feature roles during training, providing the first direct, empirical evidence of hierarchical learning in neural networks. 
    \item We provide extensive empirical validation across multiple datasets and architectures, demonstrating that our dynamic analysis is a powerful diagnostic tool for identifying insufficient model capacity and providing early-warning signals for overfitting.
\end{enumerate}

\section{Related work}
\label{sec:related_work}

FFCA builds on several key XAI research areas. Foundational feature attribution methods like LIME~\cite{ribeiro2016should} and SHAP~\cite{lundberg2017unified} provide single importance scores but can obscure complex behaviors like non-linearity or interactions, which FFCA aims to disambiguate.

Our work is closely related to methods for quantifying feature interactions. While visual tools like PDPs~\cite{friedman2001greedy} and ICE plots~\cite{goldstein2015peeking} can reveal interactions, quantitative measures like the H-statistic~\cite{friedman2008predictive} and Shapley Interactions provide model-agnostic scores. FFCA's Interaction score ($X_i$) differs by deriving this information directly from the model's geometry via the Hessian, offering an efficient, model-native measure.

The use of higher-order derivatives is an emerging XAI frontier~\cite{shahid2024second}. The Hessian matrix mathematically defines non-linearity (diagonal) and interactions (off-diagonal)~\cite{lerman2021explaining}. A key challenge is the ``zero-Hessian'' problem in ReLU networks. FFCA adopts an efficient activation-smoothing strategy, similar to Integrated Hessians~\cite{janizek2021explaining}, by temporarily replacing ReLU with Softplus during analysis. While other methods use the Hessian to compute raw interaction values~\cite{torop2023smoothhess, shahid2024second, tsang2020feature}, FFCA's novelty lies in its systematic framework that translates these geometric properties into the 4D signature and 8-archetype taxonomy for practical diagnosis.

Finally, FFCA bridges XAI with deep learning theory on network characterization~\cite{labatie2019characterizing, ba2022high}, which uses the same mathematical tools to analyze learning dynamics. FFCA systematizes these tools into an interpretable framework for the applied XAI practitioner.  Full discussion in Appendix \ref{app:related_work}.

\section{Feature-function curvature analysis (FFCA)}
\label{sec:ffca_methodology}

FFCA is a model-centric analysis framework designed for any differentiable model.
It operates by performing a geometric analysis of the model's learned function to produce a multi-dimensional characterization of each feature's role.


\subsection{Geometric intuition and problem formulation}
We consider a trained model with output $g(x)\in\mathbb{R}^k$ for input $x\in\mathbb{R}^d$ ($k=1$ for regression, $k\ge 2$ for classification). 
FFCA analyzes a \emph{scalar} scoring function $f(x)$ derived from $g(x)$ and takes all derivatives with respect to this $f$. In regression ($k=1$), We set $f(x)=g(x)$ (the scalar model output).In multiclass classification ($k\ge 2$), let $z(x)\in\mathbb{R}^k$ denote the pre-softmax logits. In our implementation, FFCA uses the \emph{maximum logit} as the scalar score, i.e., $f(x)=\max_j z_j(x)$, which corresponds to differentiating the predicted class’ logit. 
During FFCA analysis we temporarily switch ReLU activations to smooth activations (Softplus) to obtain meaningful second derivatives. 
When using the diagonal Hessian approximation, FFCA reports impact, stability, and non-linearity; cross-feature \emph{interaction} scores are computed only under the full-Hessian setting.

The core idea of FFCA is to probe the local geometry of this function's manifold using its Taylor expansion as an inspiration. The first partial derivative, $\frac{\partial f}{\partial x_i}$, represents the local, linear effect (the gradient) of feature $x_i$. Its magnitude indicates the instantaneous impact. By a first-order Taylor expansion, $f(x+\delta e_i)\approx f(x)+\frac{\partial f(x)}{\partial x_i}\delta$, so $|\partial f/\partial x_i|$ is the local Lipschitz sensitivity of $f$ to $x_i$. Using the \emph{expected} magnitude across $\mathcal{D}$ yields a robust global summary of that sensitivity.

The second partial derivatives describe the curvature. The direct second derivative, $\frac{\partial^2 f}{\partial x_i^2}$, measures the feature's own non-linearity. The mixed partial derivative, $\frac{\partial^2 f}{\partial x_i \partial x_j}$, measures the interaction between features $x_i$ and $x_j$. Second derivatives quantify curvature: $\partial^2 f/\partial x_i^2$ captures feature-wise non-linearity (convexity/concavity), while mixed partials $\partial^2 f/\partial x_i\partial x_j$ capture interactions via departures from additivity in the local quadratic approximation of $f$. This aligns with Hessian-based notions of interaction used in recent explainability work. By systematically computing and aggregating these derivatives over the data distribution, FFCA builds a comprehensive profile of each feature's behavior.

\subsection{The FFCA pipeline}
The methodology consists of three stages: input normalization (analysis-time only), derivative computation, and signature generation.

\subsubsection{Stage 1: analysis-time input normalization}
Before computing derivatives, we normalize inputs solely for FFCA analysis to make per-feature derivatives comparable across scales and distributions. All input features are scaled to have zero mean and unit variance, making the magnitudes of derivatives comparable across features with different natural scales
Concretely, we (i) z-score each feature using a scaler fit on the analysis subset, and (ii) apply a marginal rank-uniform transform to map each feature to $[0,1]$. To mitigate the influence of differing distributional shapes (e.g., skew, outliers), a quantile transformation maps each feature's values to a uniform distribution. A StandardScaler is fit on the sampled analysis data and reused for additional FFCA runs; features are then rank-mapped to $[0,1]$ per dimension. Model parameters are never updated during FFCA.
This procedure standardizes the coordinate system in which derivatives are measured; it does not alter model training or deployed inference.



\subsubsection{Stage 2: scalable derivative computation}

\paragraph{Activation smoothing.}
For networks with non-smooth activations (e.g., ReLU), second derivatives are zero almost everywhere, making curvature estimates ill-posed.
During FFCA analysis only, we replace ReLU with a \emph{smooth surrogate} (Softplus) while keeping all learned weights fixed; this enables meaningful second-order derivatives without altering the trained parameters.
Softplus is monotone, near-zero for large negative inputs, and asymptotically linear for large positive inputs, thus preserving ReLU’s qualitative behavior while smoothing the kink.
Formally, the temperatured Softplus $s_\beta(x)=\tfrac{1}{\beta}\log(1+e^{\beta x})$ converges pointwise to $\max(0,x)$ as $\beta\!\to\!\infty$.
Predictions may shift slightly during analysis; original activations are restored afterward.

\paragraph{Hessian computation: why and when.}
FFCA uses second-order information for two purposes: (i) \emph{nonlinearity} (per-feature curvature) and (ii) \emph{interaction} (cross-feature curvature).
The former only requires the Hessian \emph{diagonal}; the latter requires \emph{off-diagonal} terms.

\textbf{Diagonal Hessian (default):} Compute only $\frac{\partial^2 f}{\partial x_i^2}$; scales roughly $O(d)$ per point and suffices for the nonlinearity score.
\textbf{Full Hessian (for interactions):} Compute all entries to capture cross-feature coupling; scales roughly $O(d^2)$ per point and is used selectively when interaction analysis is required. Depending on the model and the number of features, calculating the full Hessian is computationally impossible, yet, in most cases when we deal with significant number of features like CNN based model, we can significantly improve this calculation by focusing on important parts of models attention or use optimize methods to calculate the Hessian.

While FFCA currently employs a direct and efficient activation-smoothing technique, the core interpretation framework is modular. We envision a system where alternative derivative estimation techniques, such as the Gaussian convolution approach of SmoothHess~\cite{torop2023smoothhess}, could be used as a `pluggable' backend. This would allow practitioners to select the most appropriate computational method for their specific needs---for instance, using SmoothHess's controllable covariance for highly targeted analysis---while still benefiting from FFCA's unique 4D signature and archetype interpretation.

\subsubsection{Stage 3: the 4D signature generation}
The derivatives are computed over a representative data sample, $\mathcal{D}$, and aggregated to form the final 4-dimensional signature for each feature $x_i$.
Throughout this section, we use \(\mathcal{D}\) to denote the FFCA analysis set (full eval split or a stratified subsample), not necessarily the entire training dataset.

\begin{enumerate}
    \item \textbf{Impact ($I_i$):} The expected absolute value of the first derivative.
$$I_i = \mathbb{E}_{x \in \mathcal{D}}\Bigl[\bigl|\frac{\partial f(x)}{\partial x_i}\bigr|\Bigr]$$

    \item \textbf{Volatility ($S_i$):} The variance of the first derivative.
A high score indicates a highly contextual effect.
    $$S_i = \text{Var}_{x \in \mathcal{D}}\Bigl[\frac{\partial f(x)}{\partial x_i}\Bigr]$$
    
    \item \textbf{Non-linearity ($N_i$):} The expected absolute value of the direct second derivative.
$$N_i = \mathbb{E}_{x \in \mathcal{D}}\Bigl[\bigl|\frac{\partial^2 f(x)}{\partial x_i^2}\bigr|\Bigr]$$
    
    \item \textbf{Interaction ($X_i$):} The sum of expected absolute values of the mixed partial derivatives. This score is derived from the off-diagonal elements of the Hessian matrix, as detailed below.

The Interaction score, $X_i$, provides a quantitative measure of a feature's total involvement in pairwise interactions. Its calculation is grounded in the mathematical properties of the model's Hessian matrix.
For a given feature $i$, the total interaction score is defined as the sum of the expected absolute values of all its mixed partial derivatives with other features $j$:
$$X_i = \sum_{j \neq i} \mathbb{E}_{x \in \mathcal{D}}\Bigl[\bigl|\frac{\partial^2 f(x)}{\partial x_i \partial x_j}\bigr|\Bigr]$$

where $\frac{\partial^2 f}{\partial x_i \partial x_j}$ is the cross-derivative (or mixed partial derivative) between features $x_i$ and $x_j$.
\end{enumerate}

\paragraph{Computational Process.} The calculation proceeds in three steps: \textbf{1) Compute Full Hessian Matrix:} For each data sample in a representative batch, FFCA computes the complete $d \times d$ Hessian matrix of second derivatives. \textbf{2) Extract Cross-Derivatives:} For each feature $i$, the absolute values of all off-diagonal elements in its corresponding row (or column) are summed. \textbf{3) Average Across Samples:} The interaction scores computed for each sample are then averaged across the entire dataset $\mathcal{D}$ to produce the final, stable $X_i$ value.

\paragraph{Rationale.} This approach is effective because the cross-derivative $\frac{\partial^2 f}{\partial x_i \partial x_j}$ directly measures how the gradient of the function with respect to $x_i$ changes as $x_j$ is perturbed. If this value is non-zero, it signifies that the effect of feature $i$ is dependent on the value of feature $j$---the mathematical definition of an interaction.

\section{Interpretation framework: the eight feature archetypes}
\label{sec:archetypes}

While the 4D signature provides a rich, quantitative description, its true power is unlocked through a qualitative interpretation framework. To bridge the gap from complex geometric data to actionable insights, we introduce a taxonomy of eight distinct ``feature archetypes.'' This framework is not merely descriptive; it is a prescriptive diagnostic tool designed to guide an analyst's workflow.

\subsection{Justifying the archetype space}
The FFCA signature can be conceptually divided into two distinct components: magnitude and character. \textbf{Magnitude (Impact):} This dimension answers the traditional XAI question: ``How strong is this feature's effect?'' and \textbf{Character (Volatility, Non-linearity, Interaction):} These three dimensions answer the novel question posed by FFCA: ``What is the nature of this feature's effect?''

This conceptual separation is the core of our framework. We interpret a feature's other three characteristics in respect to its Impact. The current archetype framework is purely empirical and based on what could be useful in feature role interpretation. This framing represents the set of fundamental behavioral roles a feature can play based on the geometry of the model function. We argue that eight archetypes strike the optimal balance between comprehensiveness and interpretability.

\subsection{The archetype taxonomy}

The eight archetypes are organized into three logical groups based on their complexity and role: \emph{Foundational Roles}, \emph{Complex Soloists}, and \emph{Team Players \& Complex Roles}. To provide a clear, actionable guide for practitioners, Table~\ref{tab:archetypes_8} in Appendix \ref{app:case_studies} presents a comprehensive framework summarizing each archetype. For each one, the table details its characteristic signature profile, its intuitive description, and a direct, actionable recommendation for further analysis. This structure serves as a diagnostic reference for translating FFCA signatures into practical, user-focused insights.

\section{Experimental validation}
\label{sec:validation}

To establish the validity, reliability, and utility of FFCA, we conducted a comprehensive suite of experiments. The results demonstrate FFCA's ability to provide nuanced insights where traditional methods fall short. To ground FFCA in the existing state-of-the-art, we first confirmed that its Impact score ($I_i$) captures a signal consistent with established methods. On the California Housing dataset, the Pearson correlation between FFCA's Impact scores and mean absolute SHAP values was $r = 0.96$, while the correlation with Gradient Importance was $r = 0.98$. This confirms FFCA's Impact metric captures the same core importance signal, positioning FFCA as a nuanced augmentation of existing tools. For our experiments, we use a standard Multi-Layer Perceptron (MLP) with three hidden layers and ReLU activations, which are smoothed to Softplus during analysis as described in Section~\ref{sec:ffca_methodology}. We compare FFCA against SHAP, Gradient-based importance, and Permutation Importance.

\subsection{Dynamic archetype analysis: uncovering how models learn}
While the 4D signature provides a powerful snapshot of a feature's role at the end of training, it does not reveal the process by which that role was learned. To address this, we introduce Dynamic Archetype Analysis, an extension of FFCA that captures the 4D signature at regular intervals throughout the training process. This temporal analysis transforms FFCA from a static diagnostic tool into a dynamic system for observing a model's learning behavior in real-time. By tracking the evolution of Impact, Volatility, Non-linearity, and Interaction, we can move beyond asking ``what did the model learn?'' to answering a more fundamental question: ``how does the model learn?''

This dynamic view provides unprecedented, empirical evidence for several core principles of deep learning and offers practical solutions to long-standing challenges in model development. Our comprehensive experiments validate a set of key capabilities, summarized in Appendix section in Table~\ref{tab:dynamic_ffca_insights}, demonstrating that by observing the geometric evolution of the model's function, we can diagnose architectural limitations, anticipate training problems, and gain a more profound and trustworthy understanding of the learning process itself.

\subsection{Dynamic analysis case study: synthetic credit loan}
\label{sec:dynamic_case_study}

This experiment serves as the clearest validation of the entire FFCA framework, demonstrating its ability to diagnose model failure and confirm learning success. The dataset was explicitly engineered with features representing distinct archetypes (e.g., `simple\_driver`, `base\_interactor`, `partner\_interactor`).

A low-capacity model fails to learn the problem, achieving a final $R^2$ of only 0.25. Its FFCA signatures are consequently muddled and fail to capture the ground-truth roles (see Appendix \ref{app:case_studies}, Fig. \ref{fig:credit_loan_low_capacity_summary}).

\begin{figure}[htbp]
    \centering
    \includegraphics[width=\columnwidth]{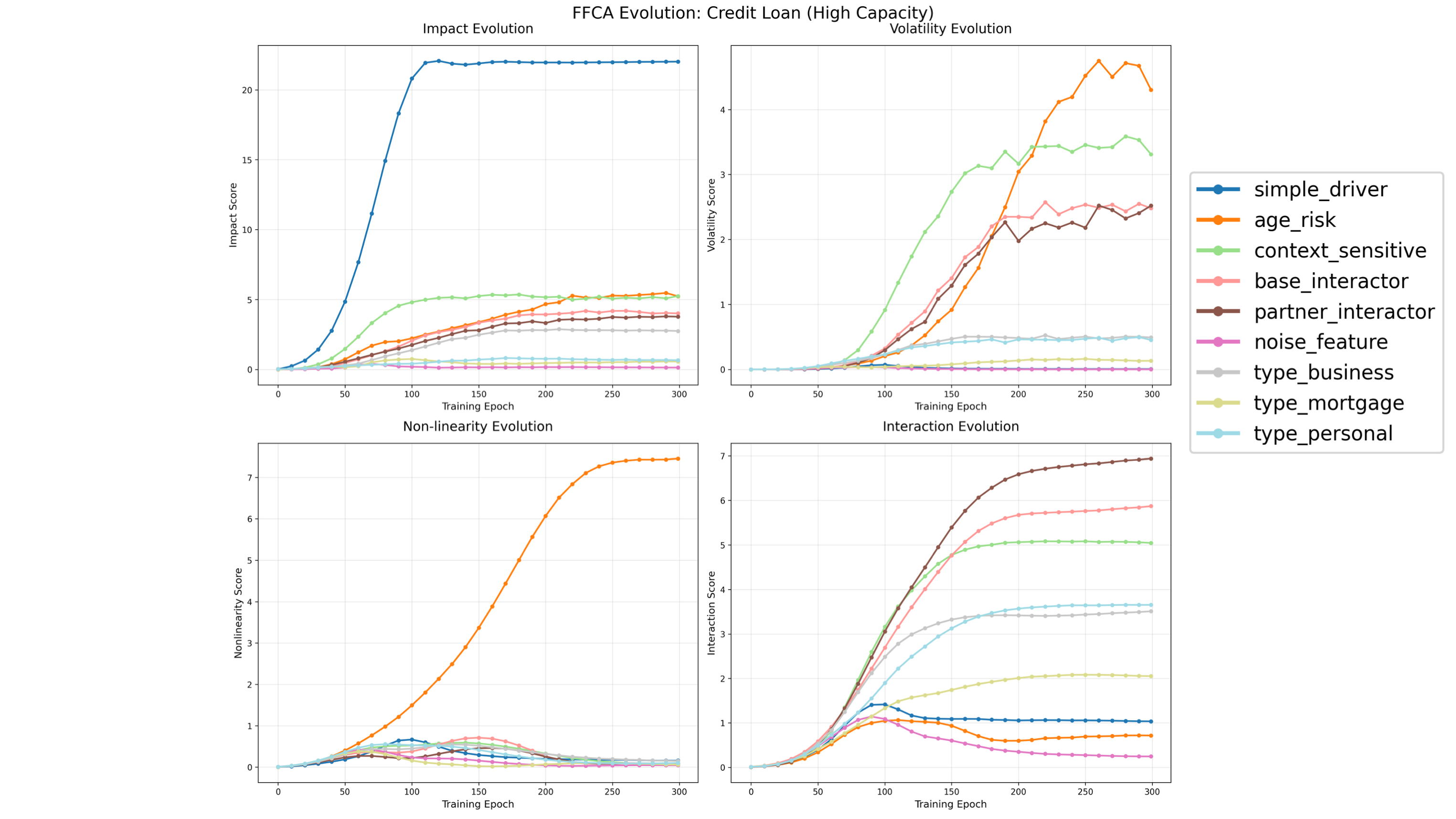}
    \caption{The learning evolution for the high-capacity model on the Credit Loan dataset. This plot clearly demonstrates hierarchical learning: the impact of the `simple\_driver' (blue line, top-left) rises and stabilizes early. In contrast, the interaction scores of the hidden interactors (red and purple lines, bottom-right) remain near zero before taking off dramatically after epoch 60, corresponding to the model's main performance improvement.}
    \label{fig:credit_loan_high_capacity_evolution}
\end{figure}

\begin{figure}[htbp]
    \centering
    \includegraphics[width=\columnwidth]{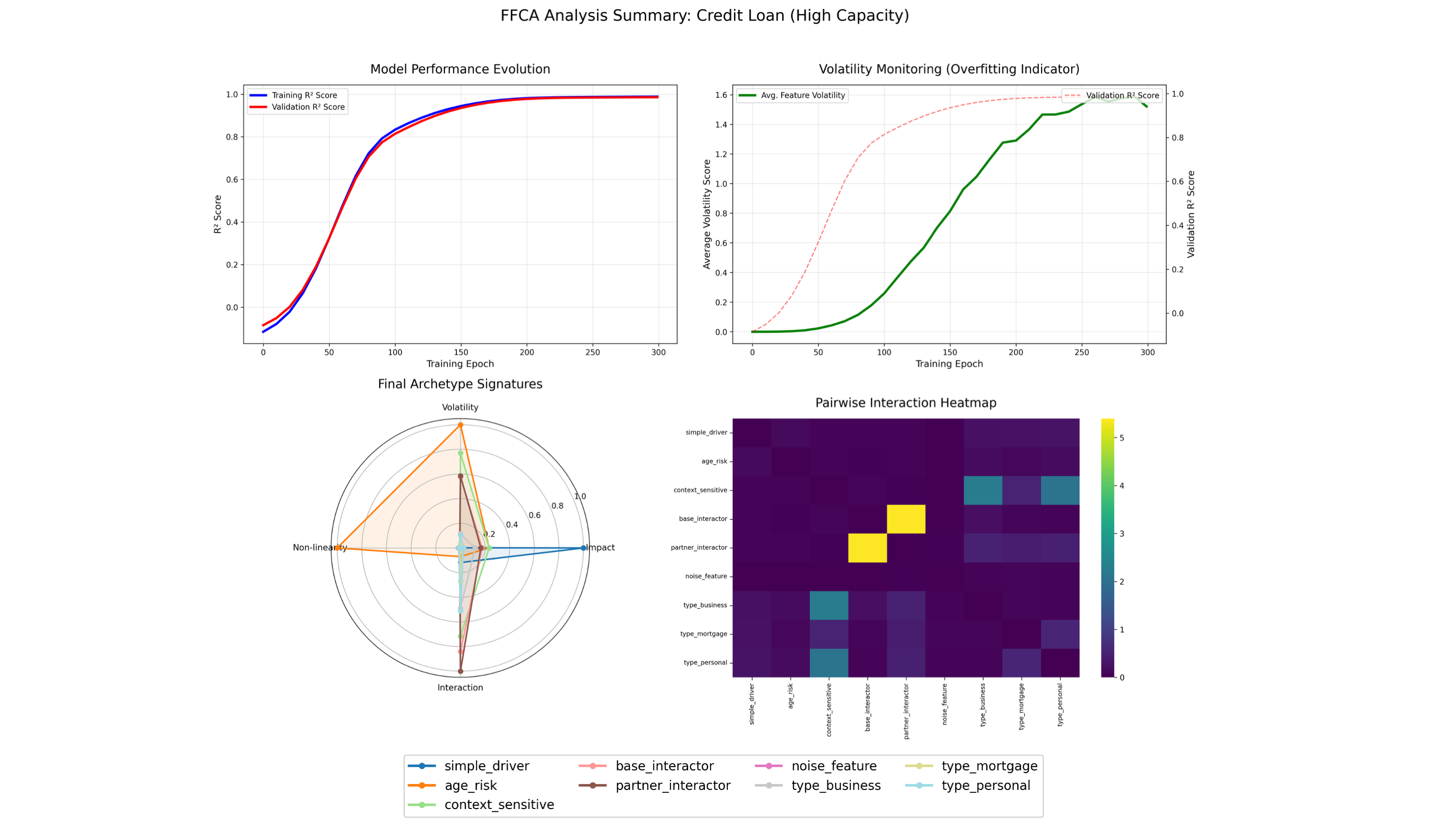}
    \caption{The final analysis summary for the high-capacity Credit Loan model. (Top-Left) The model performance reaches near-perfect $R^2$. (Top-Right) the rise and plateau of volatility reflects the successful learning of complex functions. (Bottom-Left) The final radar signatures are clear and correctly identify the engineered roles. (Bottom-Right) The interaction heatmap is showing a strong, isolated interaction between `base\_interactor' and `partner\_interactor'.}
    \label{fig:credit_loan_high_capacity_summary}
\end{figure}

In stark contrast, the high-capacity model excels, reaching an $R^2$ of 0.99 (Figure~\ref{fig:credit_loan_high_capacity_summary}). The dynamic evolution plot for this successful model (Figure~\ref{fig:credit_loan_high_capacity_evolution}) provides a canonical illustration of hierarchical learning. We observe the Impact score of the `simple\_driver' rising first, while the Interaction scores of `base\_interactor' and `partner\_interactor' remain dormant for the first 60 epochs before experiencing a dramatic ``take-off,'' which drives the model's performance to near-perfection.

The final analysis summary for the high-capacity model (Figure~\ref{fig:credit_loan_high_capacity_summary}) confirms this success. The radar plot shows distinct, unambiguous signatures for each feature. The interaction heatmap provides the ultimate proof, showing a bright square of high interaction exclusively between `base\_interactor' and `partner\_interactor', exactly as designed. All additional case studies (California Housing, Bike Sharing, Waterbirds, etc.) are detailed in Appendix \ref{app:case_studies}.
\subsection{Proof-of-ability: cross-model validation for feature engineering}
\label{sec:proof_of_ability_main}

\paragraph{Motivation.}  Beyond providing a descriptive snapshot of what a model has learned and how it learns, FFCA can offer \emph{prescriptive} insights that enable practitioners to improve future models.  Because each FFCA signature decomposes a feature’s role into interpretable components (impact, volatility, non‑linearity, and interaction), these diagnostics can suggest how to augment a simpler model so that it captures relationships that would otherwise be missed.  We conducted a controlled experiment to test whether an FFCA diagnosis from a differentiable model can guide feature engineering for a much simpler model class.

\paragraph{Experimental setup.}  We generated a challenging synthetic credit‑loan dataset whose target variable depends exclusively on non‑linear and interactive effects; there are no simple linear drivers.  A standard linear model should therefore perform poorly.  The experiment proceeded in two stages.  First, a low‑capacity MLP was trained on the dataset and analysed with FFCA; its sole purpose was to act as a differentiable ``probe'' whose 4D signatures could diagnose which features exhibited non‑linear or interaction effects.  Second, a regularised linear model (Ridge Regression) was trained on the raw features to establish a baseline.  Using the FFCA diagnosis, we engineered new features to linearise the complex effects, retrained the same linear model on the enhanced feature set, and compared performance.

\paragraph{Diagnosis.}  The FFCA analysis of the diagnostic MLP produced a clear quantitative guide, shown in Table~\ref{tab:ffca_diagnosis_values}.  The feature \texttt{dti\_ratio} was flagged as a \emph{Non‑linear Driver}: its Non‑linearity score (6.94) was more than double its Impact score (3.37).  In contrast, the features \texttt{loan\_purpose\_risk} and \texttt{num\_recent\_inquiries} were identified as \emph{Hidden Interactors}, with large Interaction scores (6.90 and 5.34) dwarfing their other signature components.  These signatures suggest that a linear model will fail because it cannot capture the curvature and synergy encoded in these features.

\begin{table}[htbp]
\centering
\caption{FFCA 4D signature obtained from the diagnostic MLP.  The high non‑linearity of \texttt{dti\_ratio} and the large interaction scores of \texttt{loan\_purpose\_risk} and \texttt{num\_recent\_inquiries} provide a concrete directive for feature engineering.}
\label{tab:ffca_diagnosis_values}
\begin{tabular}{@{}lcccc@{}}
\toprule
\textbf{Feature} & \textbf{Impact} & \textbf{Volatility} & \textbf{Non‑linearity} & \textbf{Interaction} \\
\midrule
\texttt{dti\_ratio} & 3.37 & 4.07 & \textbf{6.94} & 0.32 \\
\texttt{loan\_purpose\_risk} & 4.21 & 2.45 & 0.66 & \textbf{6.90} \\
\texttt{num\_recent\_inquiries} & 2.19 & 2.19 & 0.56 & \textbf{5.34} \\
\texttt{employment\_stability\_score} & 1.25 & 0.26 & 0.14 & 2.04 \\
\bottomrule
\end{tabular}
\end{table}

\paragraph{Action.}  The baseline Ridge Regression, trained only on the original features, achieved an $R^2$ of 0.0093—confirming that the problem was essentially unsolvable for a purely linear model.  Guided by the FFCA diagnosis, we engineered two new features: (i) a squared term $\text{dti\_ratio}^2$ to capture the high non‑linearity of \texttt{dti\_ratio}, allowing a quadratic relationship to be fit by the linear model; and (ii) a multiplicative interaction term $\text{loan\_purpose\_risk} \times \text{num\_recent\_inquiries}$ to explicitly model the synergistic effect between the two Hidden Interactors.  Retraining the same Ridge Regression on the augmented dataset dramatically improved performance to an $R^2$ of 0.6970, reducing the model’s unexplained error by roughly two‑thirds.  This dramatic improvement demonstrates that FFCA’s diagnostic signatures can be turned into actionable feature engineering directives, transforming an otherwise intractable task into one that is solvable by a simple and interpretable model.

\section{Discussion, limitations, and future work}
\label{sec:discussion}

By adopting a geometric perspective, FFCA encourages a more rigorous form of model interrogation, shifting the focus from a simple ranking of features to a qualitative understanding of their roles. The 8-archetype framework provides a much-needed vocabulary for practitioners to discuss and debug model behavior. Our work provides the first direct, empirical evidence of hierarchical learning and demonstrates that dynamic geometric analysis is a powerful tool for diagnosing model capacity issues (Sec. \ref{sec:dynamic_case_study}) and detecting overfitting (Table \ref{tab:dynamic_ffca_insights}).

\paragraph{Scope and Limitations.}
FFCA's primary requirement is a differentiable model. While our validation focused on MLPs, the principles apply to CNNs and Transformers. The primary limitation is the $O(d^2)$ cost of the full Hessian for the Interaction score, which can be mitigated by using the diagonal approximation (which is $O(d)$) or by targeting feature subsets. For non-differentiable models, FFCA could be applied to a high-fidelity differentiable surrogate.

\paragraph{Future Work.}
This work opens several promising avenues. Key priorities include extending empirical validation to architectures like Transformers and developing best practices for interpreting signatures in those contexts. The modularity of FFCA allows for integrating alternative derivative engines (e.g., SmoothHess~\cite{torop2023smoothhess}) as pluggable backends. Finally, the temporal nature of Dynamic FFCA makes it a prime candidate for production-level model monitoring. A significant shift in a feature's archetype (e.g., a Simple Workhorse becoming a Volatile Specialist) could serve as a powerful, interpretable alarm for concept drift, prompting model retraining.

\bibliographystyle{unsrtnat}        
\bibliography{references}
\newpage
\section*{NeurIPS paper checklist}
%

\begin{enumerate}

\item Do the main claims made in the abstract and introduction accurately reflect the paper's contributions and scope?
YES

\item Does the paper discuss the limitations of the work performed by the authors?
Yes


\item For crowdsourcing experiments and research with human subjects, does the paper include the full text of instructions given to participants and screenshots, if applicable, as well as details about compensation (if any)?
Not Applicable

\item Does the paper describe potential risks incurred by study participants, whether such risks were disclosed to the subjects, and whether Institutional Review Board (IRB) approvals (or an equivalent approval/review based on the requirements of your country or institution) were obtained?
Not Applicable

\end{enumerate}

\newpage
\appendix
\section*{Appendix}

\section{Extended related work}
\label{app:related_work}

FFCA is situated at the intersection of several key research areas in XAI: foundational feature attribution, the quantification of feature interactions, the use of higher-order derivatives, and the fundamental characterization of neural network behavior.

\subsection{Foundational feature attribution methods}
The dominant paradigm in post-hoc XAI is feature attribution~\cite{molnar2022interpretable}.
LIME explains individual predictions by learning a simpler, interpretable model in a local neighborhood~\cite{ribeiro2016should}.
While flexible, LIME's explanations can be unstable and sensitive to perturbation strategy, and its local linear assumption may fail for highly non-linear models~\cite{slack2020fooling}.
SHAP provides a unified framework based on Shapley values from cooperative game theory, offering desirable theoretical properties~\cite{lundberg2017unified}.
However, common implementations can be computationally expensive and assume feature independence, which can lead to misleading attributions when features are correlated~\cite{aas2021explaining, kumar2020problems}.
While these methods aim to unveil the complexity of black-box models, they can themselves be difficult to interpret, with their outputs stemming from complex internal calculations that are not always transparent to the end-user. Gradient-based methods like Integrated Gradients are computationally efficient for differentiable models but typically rely only on first-order information, failing to provide a structured decomposition of effects into components like volatility or interaction~\cite{sundararajan2017axiomatic}.

\subsection{Methods for quantifying feature interactions}
A parallel line of research focuses specifically on identifying and measuring feature interactions, a known challenge for many XAI methods~\cite{friedman2008predictive}. The importance of modeling feature interactions is well-established, particularly in domains like recommender systems where the joint effect of features is critical for accuracy.
Visual tools like Partial Dependence Plots (PDPs) and Individual Conditional Expectation (ICE) plots can reveal the presence of interactions, with non-parallel ICE lines being a strong indicator~\cite{goldstein2015peeking, friedman2001greedy}.
FFCA's Volatility and Non-linearity metrics can be seen as quantitative summaries of the phenomena these plots visualize.
To move beyond visual inspection, Friedman's H-statistic was developed as a model-agnostic measure based on the variance decomposition of partial dependence functions~\cite{friedman2008predictive}. More recently, extensions of Shapley values, known as Shapley Interactions, have been proposed to quantify higher-order interactions in a game-theoretic framework. While powerful, these methods are often model-agnostic and can be computationally intensive to approximate. In contrast, FFCA's Interaction score ($X_i$) is derived directly from the model's intrinsic geometry via the Hessian matrix, offering a more native and efficient measure of interaction for differentiable models.

\subsection{Higher-order derivatives in explainability}
The use of second-order derivatives (the Hessian matrix) is an emerging frontier in XAI, promising deeper insights than first-order methods alone~\cite{shahid2024second}. The first derivative measures slope (rate of change), but the second derivative measures curvature, or how the slope itself changes. This is fundamental for understanding non-linearity and interactions; the mixed partial derivative $\frac{\partial^2 f}{\partial x_i \partial x_j}$ directly quantifies how the effect of $x_i$ changes as $x_j$ changes, which is the mathematical definition of an interaction~\cite{lerman2021explaining}.

However, a significant practical challenge arises with modern neural networks that predominantly use the Rectified Linear Unit (ReLU) activation function. Because ReLU is piecewise-linear, its second derivative is zero almost everywhere, rendering naive Hessian-based analysis ineffective. Recent work has proposed several innovative strategies to address this ``zero-Hessian'' problem. These solutions range from mathematically sophisticated post-hoc analyses, such as \emph{SmoothHess}, which estimates the Hessian of the network function convolved with a Gaussian by leveraging Stein's Lemma~\cite{torop2023smoothhess}, to computationally intensive methods like \emph{Second Glance}, which trains an entirely separate surrogate model to predict the gradients of the primary network~\cite{shahid2024second}.

A more direct and efficient approach, which FFCA adopts, is to temporarily replace the ReLU activations with a smooth, differentiable equivalent during the analysis phase. This strategy finds strong validation in methods like \emph{Integrated Hessians}, which successfully computes axiomatic interaction values by substituting ReLU with a Softplus function post-hoc, requiring no model retraining~\cite{janizek2021explaining}. Similarly, \emph{GradientNID} employs smooth activations within a local explainer model to detect instance-specific interactions~\cite{tsang2020feature}.

While these methods provide powerful techniques for \emph{calculating} second-order effects, their primary output remains a matrix of raw interaction values. FFCA's core contribution is the comprehensive and systematic framework built upon this geometric foundation. It is novel in its systematic translation of these geometric properties into a practitioner-focused, interpretable framework---the 4D signature and its corresponding archetypes. While other toolboxes use the Hessian for reliability analysis or to compute interaction scores, FFCA transforms these geometric properties into a diagnostic system for practical model explanation.

\subsection{Characterizing learned representations in neural networks}
A parallel line of research in deep learning theory focuses on \emph{characterizing} the behavior of neural networks, often at initialization and during training~\cite{labatie2019characterizing}.
These works analyze how signals and noise propagate through layers and how meaningful feature representations are learned from data, even in the presence of noise~\cite{ba2022high, labatie2019characterizing}.
This field uses the same mathematical objects as FFCA---gradients and Hessians---to understand the network's inductive bias and whether it learns ``well-behaved'' or ``pathological'' functions.
FFCA can be seen as a bridge between these two fields.
It takes the mathematical tools from fundamental network characterization and systematizes them into an interpretable framework (the 4D signature and archetypes) for the applied XAI practitioner. This grounding suggests that FFCA is not an ad-hoc invention but a principled application of deep learning theory to the problem of interpretability, transforming the tools used to analyze loss landscapes into a diagnostic system for practical model explanation.

\section{Static analysis examples and validation}
\label{app:static_analysis}

\subsection{In-line examples for 4D signature definitions}
These examples, originally in Section \ref{sec:ffca_methodology}, demonstrate the value of the individual FFCA metrics.

\paragraph{Example for Diagnosing Volatility.}
A high Volatility score is a critical warning that a feature's effect is inconsistent across the dataset and that its global importance score may be misleading. To demonstrate this, we trained a model on a synthetic loan dataset where exist a feature named `debt\_to\_income\_ratio' that was engineered to be highly detrimental for personal loans but neutral for business loans. As shown in Table~\ref{tab:volatility_example}, a standard method like SHAP assigns a high score to `debt\_to\_income\_ratio', but cannot explain the nature of its importance. FFCA, however, correctly identifies `loan\_purpose\_Personal'---the feature representing the context switch---as having the highest Volatility (4.19). This score directly informs the analyst that the model's behavior is highly dependent on this context, prompting them to perform a sliced analysis (e.g., grouping by loan purpose) to uncover the true, nuanced relationship.

\begin{table}[htbp]
\centering
\footnotesize
\caption{FFCA's volatility score identifies the source of inconsistent feature effects. While SHAP shows `debt\_to\_income\_ratio' is important, FFCA's high Volatility for `loan\_purpose\_Personal' correctly diagnoses the model's context-dependent behavior.}
\label{tab:volatility_example}
\begin{tabular}{@{}lcccc@{}}
\toprule
\textbf{Feature} & \textbf{FFCA Impact} & \textbf{FFCA Volatility} & \textbf{SHAP} & \textbf{Permutation} \\ 
\midrule
annual\_income & 18.1347 & 0.0027 & 18.1703 & 0.6848 \\
debt\_to\_income\_ratio & 20.7971 & 2.2989 & 17.6321 & 1.1441 \\
\textbf{loan\_purpose\_Personal} & \textbf{2.2054} & \textbf{4.1854} & \textbf{6.0953} & \textbf{0.3394} \\ 
\bottomrule
\end{tabular}
\end{table}

\paragraph{Example for Uncovering Non-Linear Effects.}
A high Non-linearity score indicates that a feature's contribution is curved (e.g., exhibiting diminishing returns or a U-shaped effect), a characteristic that single attribution scores fail to capture. In our synthetic loan dataset, the `age' feature was designed to have a U-shaped relationship with risk, where both young and old applicants are penalized. Table~\ref{tab:nonlinearity_example} shows that while SHAP assigns `age' a moderate importance score, it provides no insight into the nature of this effect. In contrast, FFCA assigns `age' the highest Non-linearity score by a large margin (6.00). This immediately signals to the analyst that a simple linear interpretation is insufficient and that the feature's true behavior must be investigated with visualization tools like Partial Dependence Plots (PDPs).

\begin{table}[htbp]
\centering
\footnotesize
\caption{ FFCA's non-linearity score flags features with curved relationships. While SHAP gives `age' a moderate score, FFCA's high Non-linearity score provides the crucial, actionable insight that the feature's effect is not linear.}
\label{tab:nonlinearity_example}
\begin{tabular}{@{}lcccc@{}}
\toprule
\textbf{Feature} & \textbf{FFCA Impact} & \textbf{FFCA Non-linearity} & \textbf{SHAP} & \textbf{Permutation} \\ 
\midrule
annual\_income & 18.1347 & 0.0998 & 18.1703 & 0.6848 \\
\textbf{age} & \textbf{4.3211} & \textbf{5.9986} & \textbf{4.4595} & \textbf{0.0524} \\
loan\_purpose\_impact & 6.8591 & 0.0622 & 6.2144 & 0.0869 \\ 
\bottomrule
\end{tabular}
\end{table}

\paragraph{Example for detecting Interaction.} This example designed to unmask Hidden Interactors, showcasing FFCA's unique ability to resolve ambiguity where single-score methods are confounded. We conduct a two-part ablation study using a synthetic ``Deceptive Loan'' dataset.

Setup: Two-World Comparison.
To create an unambiguous test, we generate two distinct datasets representing two self-consistent ``worlds'':

\begin{enumerate}
    \item \textbf{World 1 (Full Model):} A dataset is generated where the target variable is strongly influenced by a multiplicative interaction between two features: `credit\_history\_len' and `num\_recent\_inquiries'. These features have negligible individual linear effects. The model also includes a strong linear driver (`annual\_income'), a moderate linear driver (`loan\_purpose\_impact'), and a noise feature.
    \item \textbf{World 2 (Ablated Model):} A second dataset is generated from scratch where the `num\_recent\_inquiries' feature and its corresponding interaction term \emph{never existed}. The target variable is determined only by the direct, linear effects of the remaining features.
\end{enumerate}

We then train a separate MLP model on each dataset and run a comparative XAI analysis. This rigorous setup allows us to directly observe how the measured importance of `credit\_history\_len' changes when its interaction partner is fundamentally absent.

\paragraph{Results: Disambiguating Feature Roles.}
The results, summarized in Table~\ref{tab:deceptive_ablation}, provide a clear and powerful demonstration of FFCA's diagnostic capabilities. In the full model (World 1), both SHAP and FFCA's Impact score assign high importance to `credit\_history\_len' (7.56 and 7.01, respectively), making it appear similar to the genuine linear driver `loan\_purpose\_impact'. However, only FFCA's high Interaction score (2.45) reveals the true, underlying reason for its importance.

\begin{figure}[htbp]
    \centering
    \includegraphics[width=\columnwidth]{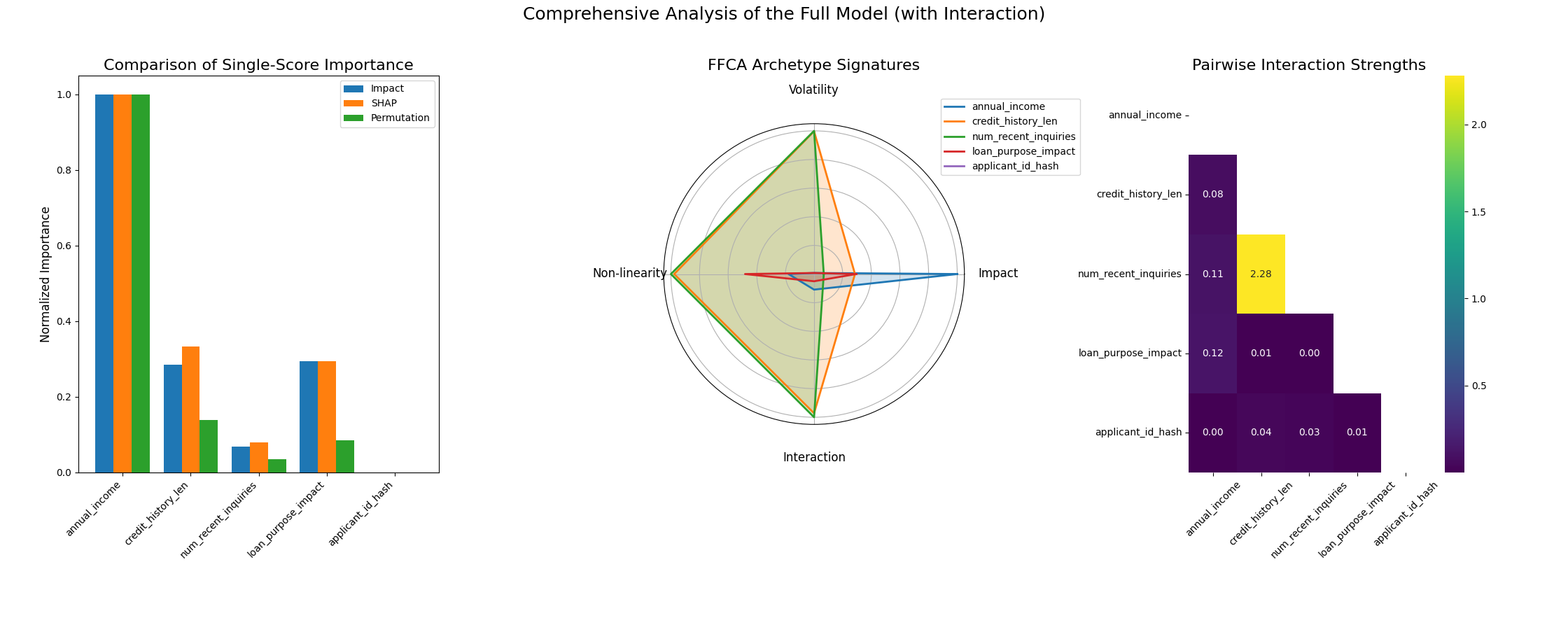}
    \caption{Comprehensive analysis of the full model with the interaction term. \textbf{(Left)} Single-score methods ambiguously rank `credit\_history\_len' and `loan\_purpose\_impact'. \textbf{(Center)} The FFCA radar plot clearly distinguishes their archetypes: `credit\_history\_len' is dominated by Interaction (orange), while `loan\_purpose\_impact' is driven by Impact (green). \textbf{(Right)} The interaction heatmap confirms the strong pairwise relationship between the two hidden interactors.}
    \label{fig:deceptive_full_analysis}
\end{figure}

\begin{figure}[htbp]
    \centering
    \includegraphics[width= 10.0cm]{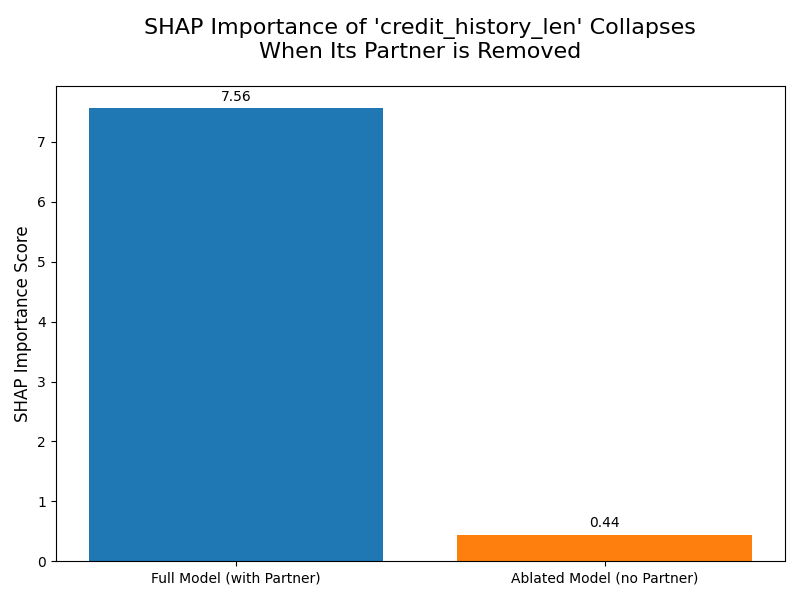}
    \caption{The SHAP importance of `credit\_history\_len' collapses from 7.56 to 0.44 when its interaction partner is removed from the system, proving its importance was derived from the interaction that FFCA identified.}
    \label{fig:deceptive_ablation_plot}
\end{figure}

The ablation study confirms this diagnosis. When the model is retrained in World 2 without the interaction partner, the SHAP importance of `credit\_history\_len' collapses from 7.56 to 0.44---a drop of over 94\%. Its FFCA Impact score likewise plummets from 7.01 to 0.31. This experiment proves that the importance attributed to `credit\_history\_len' by single-score methods was almost entirely an artifact of an interaction that only FFCA could explicitly identify and quantify. The comprehensive analysis of the full model is visualized in Figure~\ref{fig:deceptive_full_analysis}, and the dramatic collapse in importance is shown in Figure~\ref{fig:deceptive_ablation_plot}.

\begin{table}[htbp]
\centering
\footnotesize
\caption{ Comparative results from the two-part ablation study. In the full model, SHAP and FFCA impact scores for `credit\_history\_len' are misleadingly high. After removing its interaction partner (`num\_recent\_inquiries'), its importance collapses, a behavior predicted only by FFCA's high Interaction score in the first experiment.}
\label{tab:deceptive_ablation}
\begin{tabular}{@{}llccccc@{}}
\toprule
\textbf{Experiment} & \textbf{Feature} & \textbf{FFCA Impact} & \textbf{FFCA Interaction} & \textbf{SHAP} & \textbf{Permutation} \\ 
\midrule
\multirow{5}{*}{\textbf{Full Model}} & annual\_income & 24.1827 & 0.3352 & 22.5300 & 1.6163 \\
 & \textbf{credit\_history\_len} & \textbf{7.0066} & \textbf{2.4485} & \textbf{7.5595} & \textbf{0.2233} \\
 & num\_recent\_inquiries & 1.7652 & 2.5167 & 1.8861 & 0.0558 \\
 & loan\_purpose\_impact & 7.2017 & 0.1921 & 6.6852 & 0.1378 \\
 & applicant\_id\_hash & 0.1322 & 0.0688 & 0.0899 & 0.0001 \\ 
\midrule
\multirow{4}{*}{\textbf{Ablated Model}} & annual\_income & 23.6511 & 0.4097 & 22.4743 & 1.7911 \\
 & \textbf{credit\_history\_len} & \textbf{0.3058} & \textbf{0.1769} & \textbf{0.4443} & \textbf{0.0007} \\
 & loan\_purpose\_impact & 6.7542 & 0.2870 & 6.9064 & 0.1588 \\
 & applicant\_id\_hash & 0.1393 & 0.0388 & 0.0560 & 0.0000 \\ 
\bottomrule
\end{tabular}
\end{table}

\subsection{Ground-truth validation: visualizing the ideal archetypes}
\label{app:static_archetypes}
To establish an unambiguous baseline for the FFCA signatures and validate the analysis methodology itself, we first created a controlled, deterministic dataset. The primary goal of this experiment is to isolate the FFCA measurement process from the stochasticity and unpredictability inherent in neural network training. Instead of training a model and then analyzing it, we apply FFCA directly to a custom-built mathematical function where the geometric properties corresponding to each feature are explicitly engineered. This allows us to generate the ideal or ``ground-truth'' signature for each of the eight archetypes.

\paragraph{Setup: a deterministic function.}
We scripted a GroundTruthModel, which acts as a deterministic function rather than a trained model. This function takes an 8-dimensional input vector, where each feature is designed to perfectly embody one of the eight archetypes from Table~\ref{tab:archetypes_8}. The output of the function is a direct, calculated result of specific mathematical operations chosen to produce predictable first and second-order derivatives. For instance:

\begin{itemize}
    \item \textbf{The Simple Workhorse} feature contributes to the output via a single, large linear term ($y = \dots + 30.0 \cdot x_{\text{workhorse}}$).
    \item \textbf{The Non-linear Driver} is defined by a strong quadratic term ($y = \dots + 15.0 \cdot x_{\text{nonlinear}}^2$).
    \item \textbf{The Volatile Specialist}'s contribution is multiplied by a random contextual variable ($y = \dots + 8.0 \cdot x_{\text{volatile}} \cdot \text{context}$).
    \item \textbf{The Hidden Interactor} has a deliberately tiny linear coefficient ($0.1 \cdot x_{\text{hidden}}$) but is included in large multiplicative terms ($y = \dots + 8.0 \cdot x_{\text{hidden}} \cdot x_{\text{catalyst}}$).
\end{itemize}

\begin{figure}[t]
    \centering
    \includegraphics[width=0.92\linewidth]{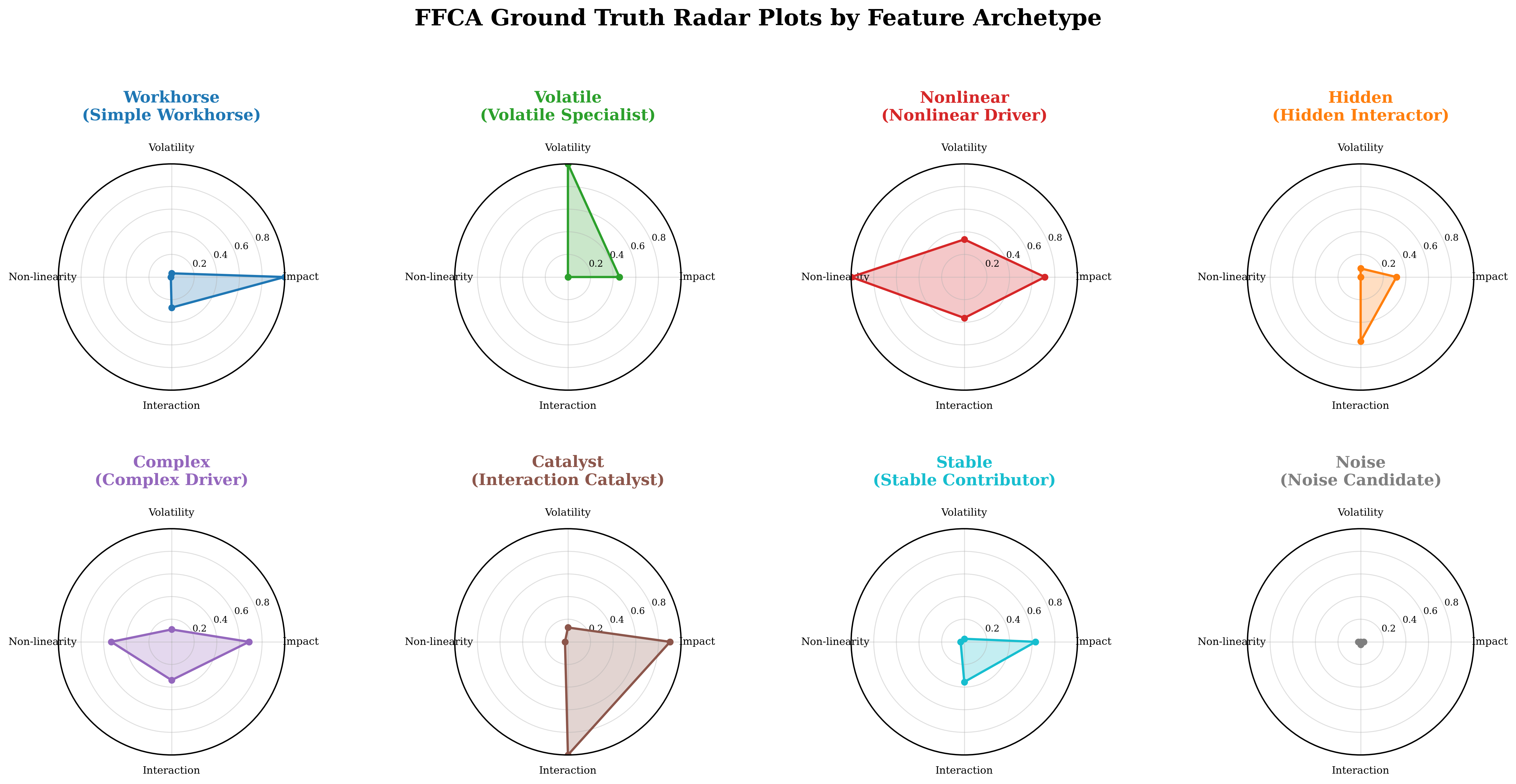}
    \caption{ The ideal FFCA signatures for the eight ground-truth archetypes. These plots are generated not from a trained neural network, but from a deterministic mathematical function in which each feature was explicitly engineered to be a perfect exemplar of its archetype. Each radar plot shows the 4D signature for one such feature, visually demonstrating the pure ``fingerprint'' for each role (e.g., the Simple Workhorse is dominated by Impact, while the Hidden Interactor is defined by Interaction). This provides a clear validation of the FFCA measurement methodology.}
    \label{fig:ground_truth_archetypes}
\end{figure}

\paragraph{Results: the archetype ``Fingerprints.''}
Applying FFCA to this deterministic function yields the ideal visual ``fingerprints'' for each archetype, shown in Figure~\ref{fig:ground_truth_archetypes}. The resulting radar plots provide a clear, visual dictionary for interpreting FFCA signatures in real-world scenarios.

\begin{table}[htbp]
\centering
\caption{The eight FFCA feature archetypes. This framework translates the 4D signature into a practical diagnostic tool, guiding subsequent analysis by providing a core description and an actionable recommendation for each feature role.}
\label{tab:archetypes_8}
{
\begin{tabular}{@{}p{0.2\textwidth} p{0.2\textwidth} p{0.28\textwidth} p{0.28\textwidth}@{}}
\toprule
\textbf{Archetype} & \textbf{Signature Profile} & \textbf{Description \& Intuition} & \textbf{Actionable Recommendation} \\ 
\midrule
\multicolumn{4}{l}{\emph{Foundational Roles: Direct and simple contributors}} \\
\addlinespace
1. Simple Workhorse & I: High; S/N/X: Low & A powerful, reliable, and direct contributor. Its effect is strong and consistent, like `annual\_income' in a loan model. & Trust global explanations. Use Partial Dependence Plots (PDPs) to confirm its linear, stable behavior. \\
\addlinespace
2. Stable Contributor & I: Mid-range; S/N/X: Low & A reliable secondary feature with a consistent but less powerful effect. Its behavior is predictable and adds value. & Acknowledge as a secondary driver. It can be trusted but is not the primary focus for explanation or feature engineering. \\
\addlinespace
3. Noise Candidate & I/S/N/X: Low & A feature with no meaningful contribution. The model has correctly learned to ignore it, like an `applicant\_id\_hash'. & Consider for feature selection. This feature adds complexity without providing signal and can likely be removed. \\
\midrule
\multicolumn{4}{l}{\emph{Complex Soloists: Important but with non-obvious, individual behavior}} \\
\addlinespace
4. Non-linear Driver & N: High; I: Mid-High & Its effect is strong but curved, exhibiting behaviors like diminishing returns or U-shaped risk (e.g., `age' in a risk model). & Visualize its 1D effect. Do not trust a linear interpretation. Use 1D PDP/ICE plots to understand its curved relationship. \\
\addlinespace
5. Volatile Specialist & S: High; I: Mid-High & A powerful but context-dependent feature whose effect changes significantly depending on other features' values. & Perform sliced analysis. Do not analyze globally. Group data by suspected context-switchers to uncover distinct local behaviors. \\
\midrule
\multicolumn{4}{l}{\emph{Team Players \& Complex Roles: Importance derived from interactions}} \\
\addlinespace
6. Hidden Interactor & X: High; I: Low & Has little direct effect but becomes critically important when combined with specific partners. Its value is synergistic. & Stop analyzing in isolation. Use 2D PDPs or SHAP interaction plots to identify its partners and understand their joint effect. \\
\addlinespace
7. Interactive Catalyst & I: High; X: High & A ``team captain'' that has a strong direct impact AND serves as a hub of interactions, amplifying other features. & Analyze both direct and joint effects. Use 1D PDPs for its individual role and 2D PDPs for its key interactions. \\
\addlinespace
8. Complex Driver & High on I and two or more of S, N, X & A multifaceted feature that is a powerful driver in multiple complex ways simultaneously (e.g., non-linear and volatile). & Requires a full deep-dive. No single explanation will suffice. Use the complete XAI toolkit (local, global, sliced, and interaction analysis). \\
\bottomrule
\end{tabular}
}
\end{table}

\section{Additional dynamic analysis case studies}
\label{app:case_studies}

This section contains the full results for all experimental case studies.

\subsection{Ground-truth validation: synthetic credit loan (Low-Capacity Model)}
\begin{figure}[htbp]
    \centering
    \includegraphics[width=\columnwidth]{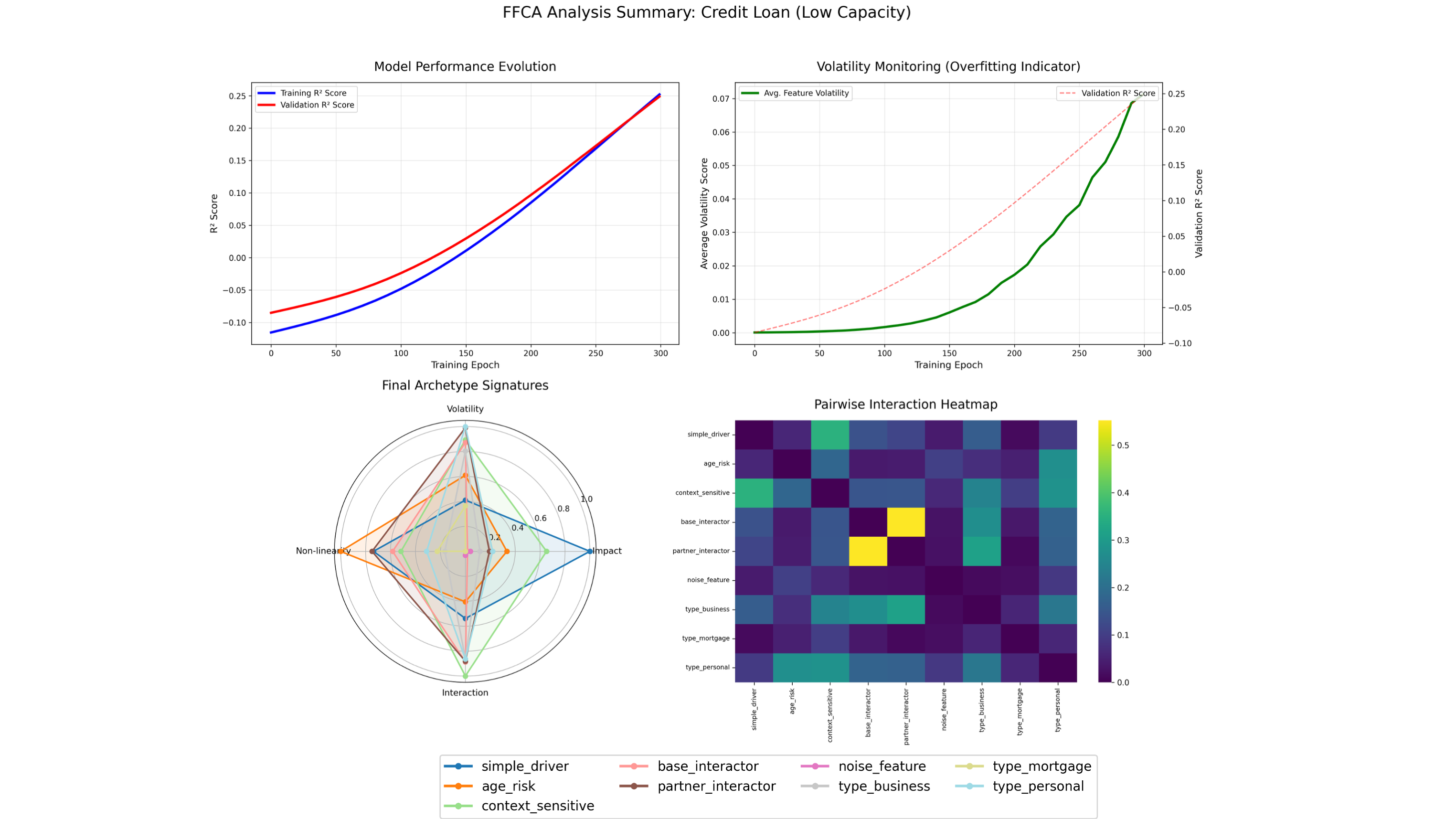}
    \caption{The final analysis summary for the low-capacity Credit Loan model. This plot illustrates a model with insufficient capacity. (Top-Left) The model fails to learn, achieving a poor $R^2$ of 0.25. (Bottom-Left \& Bottom-Right) The FFCA signatures are consequently muddled, with low absolute scores and an inability to clearly distinguish the engineered feature roles.}
    \label{fig:credit_loan_low_capacity_summary}
\end{figure}

This is the supplemental result for the case study in Section \ref{sec:dynamic_case_study}. Figure \ref{fig:credit_loan_low_capacity_summary} shows the results of training a model with insufficient capacity on the ground-truth dataset. The model fails to learn the problem, achieving a final $R^2$ of only 0.25. Its FFCA signatures are consequently muddled and fail to capture the ground-truth roles, standing in stark contrast to the successful high-capacity model shown in the main paper.

\subsection{Real-world case study: California housing}
\begin{figure}[htbp]
    \centering
    \includegraphics[width=\columnwidth]{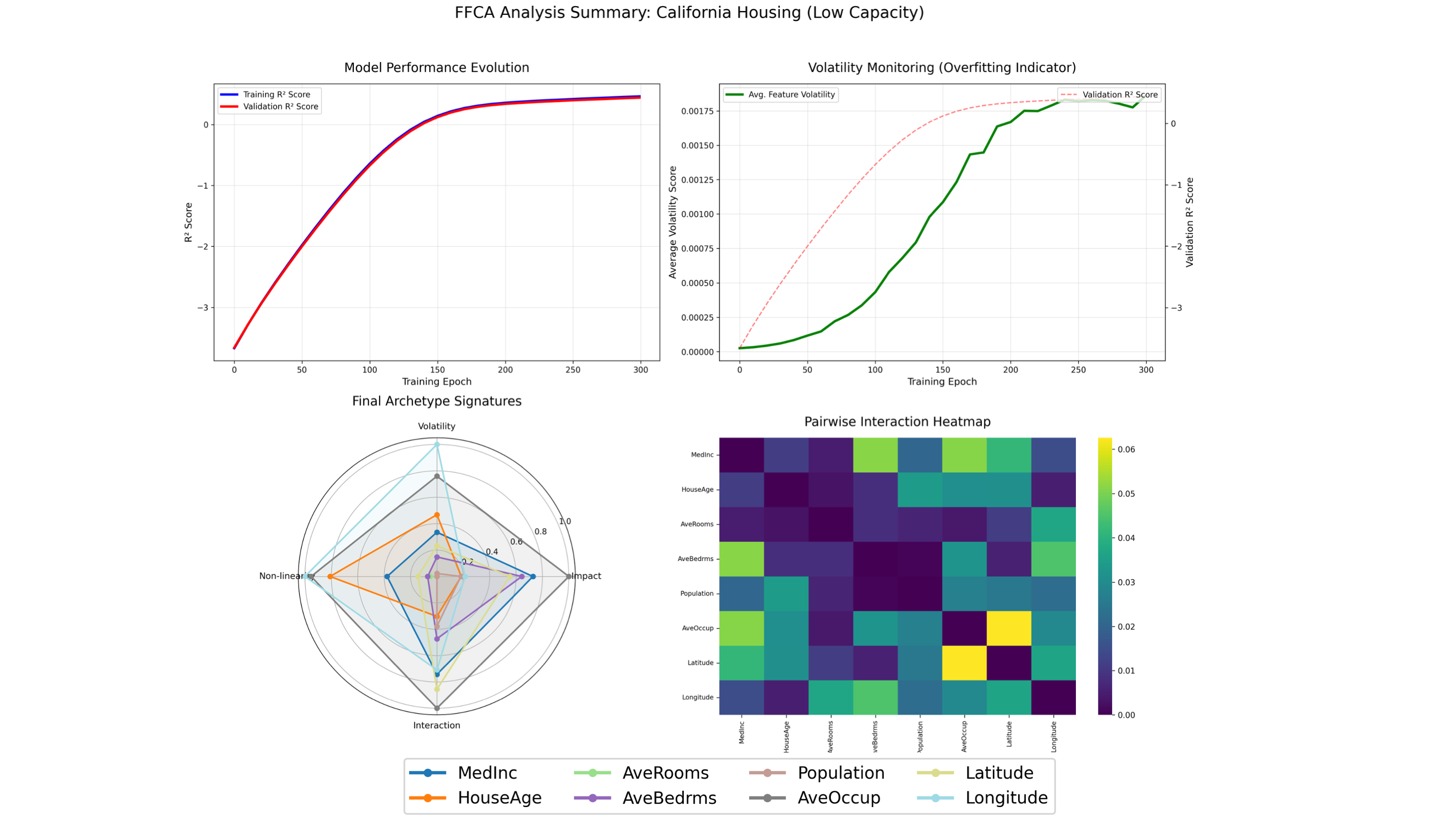}
    \caption{ Final analysis for the low-capacity California Housing model. The model achieves a modest $R^2$ of 0.44, and the FFCA signatures, while identifying some key features, are less distinct than in the high-capacity version.}
    \label{fig:california_housing_low_capacity_summary}
\end{figure}

\begin{figure}[htbp]
    \centering
    \includegraphics[width=\columnwidth]{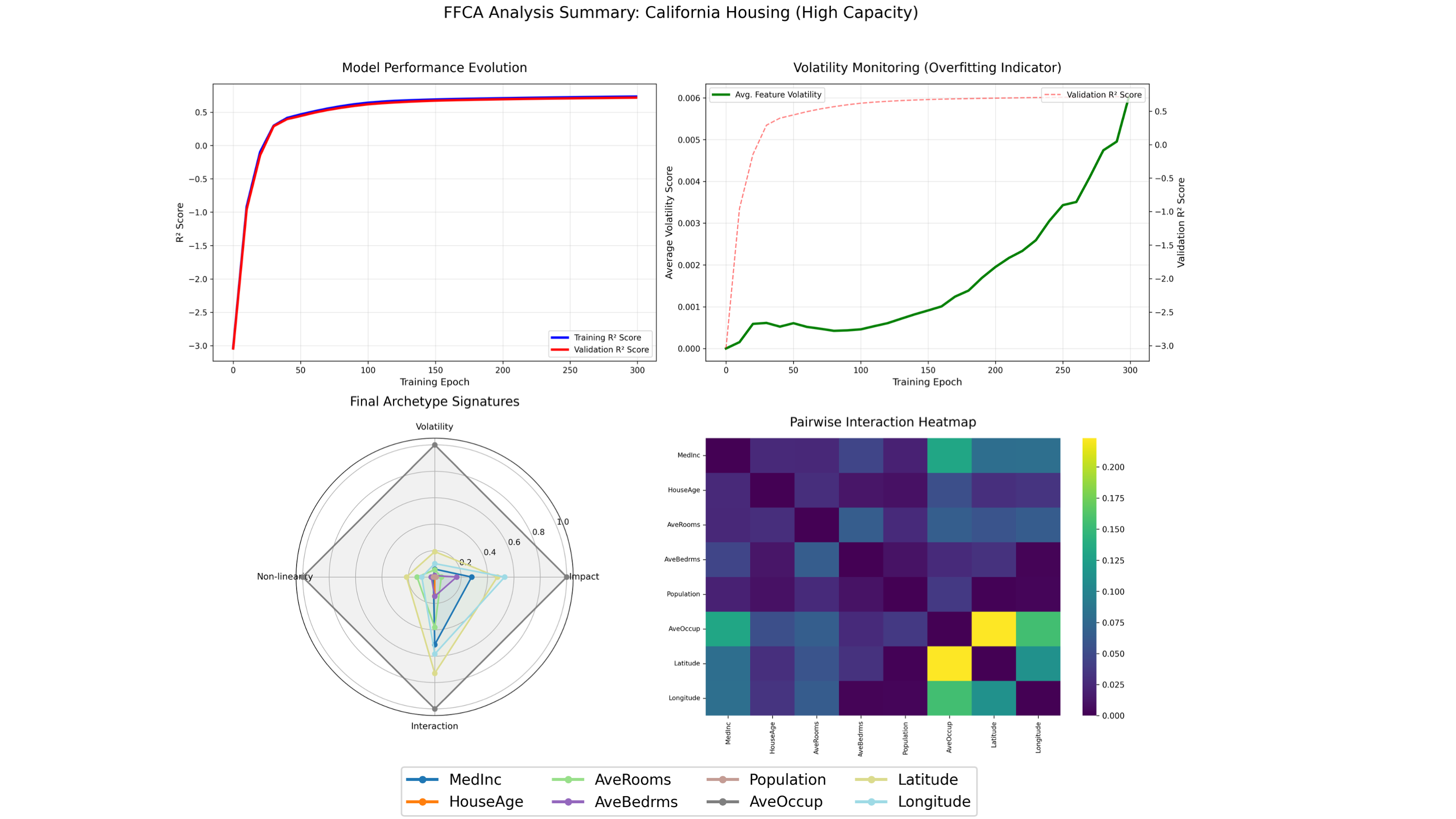}
    \caption{Final analysis for the high-capacity California Housing model. The interaction heatmap (bottom-right) is key, revealing the strong interplay between `Latitude' and `Longitude', and their joint effect with `MedInc'. The analysis also uncovers the profound role of `AveOccup' (Average Occupancy) as a Complex Driver. The interaction heatmap (bottom-right) reveals its strong interplay with `Latitude' and `Longitude', confirming the model learned that the effect of household size is highly location-dependent. The feature's complex radar plot (bottom-left) visualizes its multifaceted nature. This confirms the model learned a sophisticated pricing function.}
    \label{fig:california_housing_high_capacity_summary}
\end{figure}

On the California Housing dataset, Dynamic FFCA reveals how the model learns complex socio-geographic relationships. While the low-capacity model struggles to find clear patterns (Figure~\ref{fig:california_housing_low_capacity_summary}), the high-capacity model (Figure~\ref{fig:california_housing_high_capacity_summary}) uncovers a much richer story. While `MedInc' (Median Income) is an important driver, the most profound insight comes from AveOccup (Average Occupancy). The analysis of the high-capacity model (Figure~\ref{fig:california_housing_high_capacity_summary}) identifies AveOccup as a quintessential Complex Driver. Its radar plot is stretched across all four dimensions, indicating that it has high direct impact, its effect is non-linear, and it is highly dependent on context. It has high Impact (it's an important factor), it has high Volatility and Interaction (its effect is highly dependent on location), it has high Non-linearity (the negative effect of adding a 5th person to a house is likely much greater than the effect of adding a 3rd person).

The interaction heatmap confirms this, showing that `Latitude' and `Longitude' not only interact strongly with each other but also with `MedInc'. This demonstrates FFCA's ability to move beyond simple feature importance rankings to uncover the true, conditional nature of feature effects in a real-world setting. Furthermore, the interaction heatmap provides the explanation for AveOccup exhibits a very strong interaction with the geographic features Latitude and Longitude. This is a remarkable finding, as it shows the model learned that the significance of household size is not uniform across the state; rather, its effect on price is intensely local. A high occupancy might be standard in one neighborhood but a strong negative predictor in another, a nuanced reality of housing markets that single-score methods would completely miss. This discovery of AveOccup as a geographically-sensitive, complex feature demonstrates FFCA's power to uncover unexpected, domain-relevant insights.

\subsection{Real-world case study: bike sharing demand}
The Bike Sharing dataset, which contains temporal features, provides an excellent test for the overfitting detection hypothesis.Crucially, the Volatility Monitoring plot (top-right) provides an early warning. The average feature volatility (green line) remains stable during the initial learning phase but begins a sharp, sustained climb around epoch 100-120, precisely at the inflection point where the model's ability to generalize ceases. This spike acts as an alarm bell, signaling the onset of overfitting as the model's gradients become unstable trying to fit noise. The subsequent decline in volatility does not indicate a recovery; rather, it shows the model has finished overfitting and converged to a stable but suboptimal state, having memorized the training data's noise. This demonstrates that monitoring for the initial spike in volatility can serve as a powerful, real-time indicator to stop training, a capability that traditional validation metrics alone do not provide. The final interaction heatmap also reveals the expected strong relationships between `temp' and `atemp' (temperature and ``feels-like'' temperature) and the critical role of the `hour' feature.

\begin{figure}[htbp]
    \centering
    \includegraphics[width=\columnwidth]{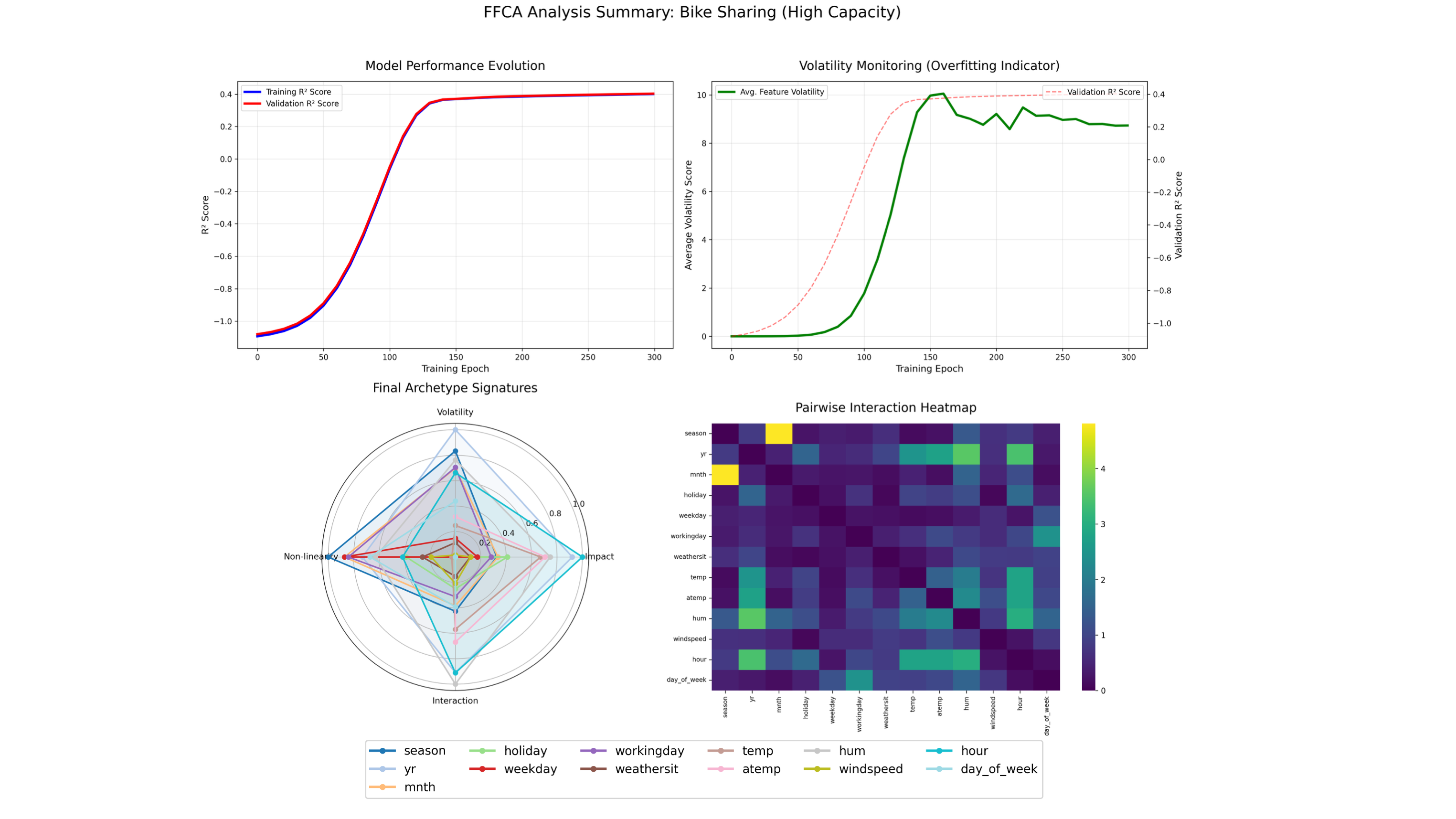}
    \caption{Final analysis for the Bike Sharing dataset. The volatility monitoring plot (top-right) shows a clear spike in average volatility (green line) that coincides with the plateau in validation performance (red dotted line). This spike is the key early warning signal for overfitting. The later decline in volatility indicates the model has stabilized in an overfitted state.}
    \label{fig:bike_sharing_high_capacity_summary}
\end{figure}

\begin{figure}[htbp]
    \centering
    \includegraphics[width=\columnwidth]{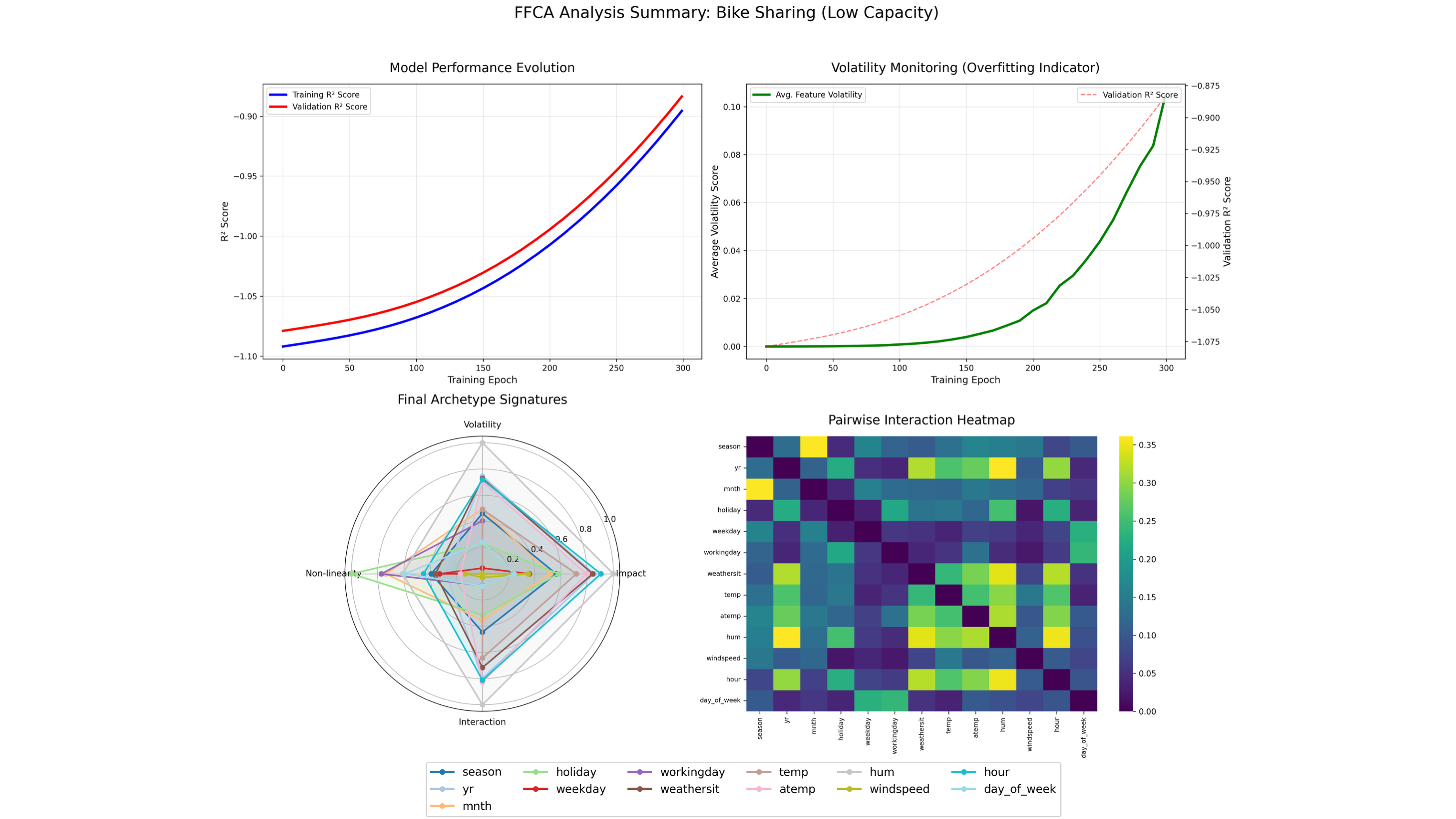}
    \caption{Final analysis for the low-capacity Bike Sharing model. The model fails to learn the task, achieving a negative final $R^2$ score. This result underscores the dataset's complexity and provides a baseline for the high-capacity model's performance.}
    \label{fig:bike_sharing_low_capacity_summary}
\end{figure}

For completeness, the analysis of the low-capacity model is presented in Figure~\ref{fig:bike_sharing_low_capacity_summary}. It fails to achieve a positive R-squared value, reinforcing the complexity of the dataset and serving as a performance baseline that highlights the partial success and clearer overfitting dynamics of the larger architecture.

\begin{table}[htbp]
\centering
\caption{Empirically validated insights from dynamic archetype analysis. This table summarizes the key research questions that can be definitively answered by observing the temporal evolution of FFCA signatures, with supporting evidence from the multi-dataset experiments.}
\label{tab:dynamic_ffca_insights}
{\footnotesize
\begin{tabular}{@{}p{0.25\textwidth} p{0.4\textwidth} p{0.3\textwidth}@{}}
\toprule
\textbf{Research Question} & \textbf{Empirical Finding \& Evidence} & \textbf{Practitioner's Guide} \\ 
\midrule
\multicolumn{3}{l}{\emph{Fundamental Learning Theory}} \\
\addlinespace
How do neural networks build complex representations? & Hierarchical Learning Confirmed: Models consistently learn in stages. Simple, linear effects (Impact) are learned first, followed by non-linearities, and finally, complex interactions. Evidence: In the Credit Loan experiment (Fig.~\ref{fig:credit_loan_high_capacity_evolution}), the Impact of `simple\_driver' rises and stabilizes early, while the Interaction scores for `base\_interactor' and `partner\_interactor' remain near zero before experiencing explosive growth in later epochs, coinciding with major performance gains. & Trust that early training epochs are dedicated to capturing the ``easy'' patterns. Do not expect complex interactions to be learned until the model has mastered the basic linear and non-linear relationships. \\
\addlinespace
\midrule
\multicolumn{3}{l}{\emph{Model Architecture \& Debugging}} \\
\addlinespace
How can we detect if a model has insufficient capacity for the data's complexity? & Complexity Score Plateaus: A model with insufficient capacity will fail to develop high Non-linearity or Interaction scores, even if those relationships exist in the data. Its performance will plateau at a suboptimal level. Evidence: The low-capacity model on the Credit Loan data achieved an $R^2$ of only 0.25, with its Interaction scores failing to develop. In contrast, the high-capacity model achieved an $R^2$ of 0.99, driven by a >10x increase in learned Interaction scores (as seen in Fig.~\ref{fig:credit_loan_high_capacity_summary}). & If model performance is poor, inspect the FFCA evolution plots. If Non-linearity and Interaction scores are flat, the model architecture is likely too simple. Increase model depth or width to provide the necessary capacity. \\
\addlinespace
Is my model learning the patterns I expect? & Ground-Truth Validation: Dynamic FFCA correctly identifies the emergence of engineered feature roles, confirming that the model learns what it is supposed to.Evidence: The high-capacity Credit Loan model correctly identified all engineered archetypes, and the interaction heatmap (Fig.~\ref{fig:credit_loan_high_capacity_summary}) shows a bright square precisely at the intersection of the two engineered interactors (`base\_interactor`, `partner\_interactor`), providing a definitive proof & Use Dynamic FFCA on a small, controlled slice of your data with known patterns to validate that your model architecture is capable of learning the kinds of relationships you expect to see in the full dataset. \\
\addlinespace
\midrule
\multicolumn{3}{l}{\emph{Training Optimization}} \\
\addlinespace
Can we predict overfitting before validation performance degrades? & Volatility as an Early Warning: A sharp, sustained increase in average feature Volatility often precedes or coincides with the point where validation performance plateaus or degrades. Evidence: In the Bike Sharing experiment (Appendix \ref{app:case_studies}, Fig. \ref{fig:bike_sharing_high_capacity_summary}), a distinct ``elbow'' in the average volatility curve appears around epoch 120, the same point where the validation $R^2$ score stops improving, indicating the model has begun to learn spurious correlations. & Monitor the average feature volatility during training. A sudden spike is a strong signal to consider early stopping, reduce the learning rate, or apply stronger regularization, even before the validation loss clearly indicates overfitting. \\
\bottomrule
\end{tabular}
}
\end{table}

\subsection{Diagnosing Shortcut Learning in CNNs: The Waterbirds Case Study}
To demonstrate FFCA's applicability beyond tabular data and to address the critical challenge of shortcut learning in computer vision, we conducted an experiment on the Waterbirds dataset. This canonical dataset is designed to induce models to learn spurious correlations. It contains images of two bird classes (landbirds and waterbirds) on two background types (land and water). The dataset is intentionally biased: most images show birds in their typical environment (e.g., waterbirds on water), creating a strong correlation between the background and the class label. The minority groups (e.g., a waterbird on land) require the model to learn the actual features of the bird itself, rather than relying on the "easy" background shortcut.

For this task, we trained a simple Convolutional Neural Network (CNN) and applied Dynamic FFCA to track its learning process. The analysis provides the definitive conclusion for shortcut learning, revealing a progressive shift in the model's attention from the bird to the background. This phenomenon is quantified by two key metrics derived from the FFCA Impact maps: the \textbf{Foreground-Background Ratio (FBR)}, which measures the ratio of impact in a central crop versus the periphery, and the \textbf{Center of Mass (CoM) Distance}, which tracks how far the model's attention is from the image center.

\begin{figure}[htbp]
    \centering
    \begin{minipage}[b]{0.98\textwidth}
        \centering
        \includegraphics[width=\textwidth]{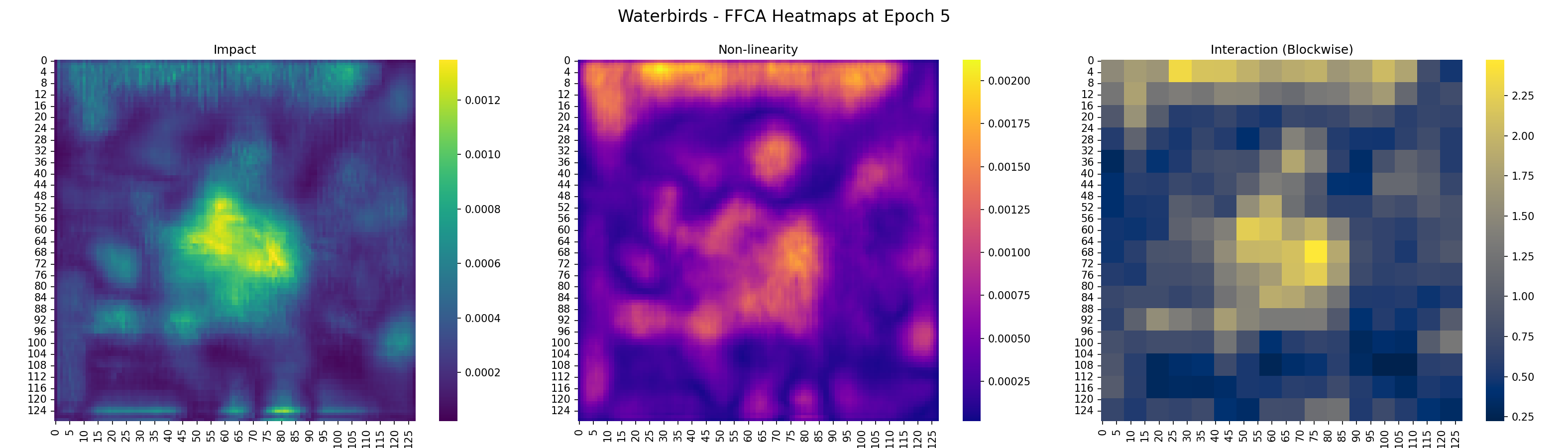}
        \caption{Epoch 5: Attention is focused on the bird.}
    \end{minipage}
    \hfill
    \begin{minipage}[b]{0.98\textwidth}
        \centering
        \includegraphics[width=\textwidth]{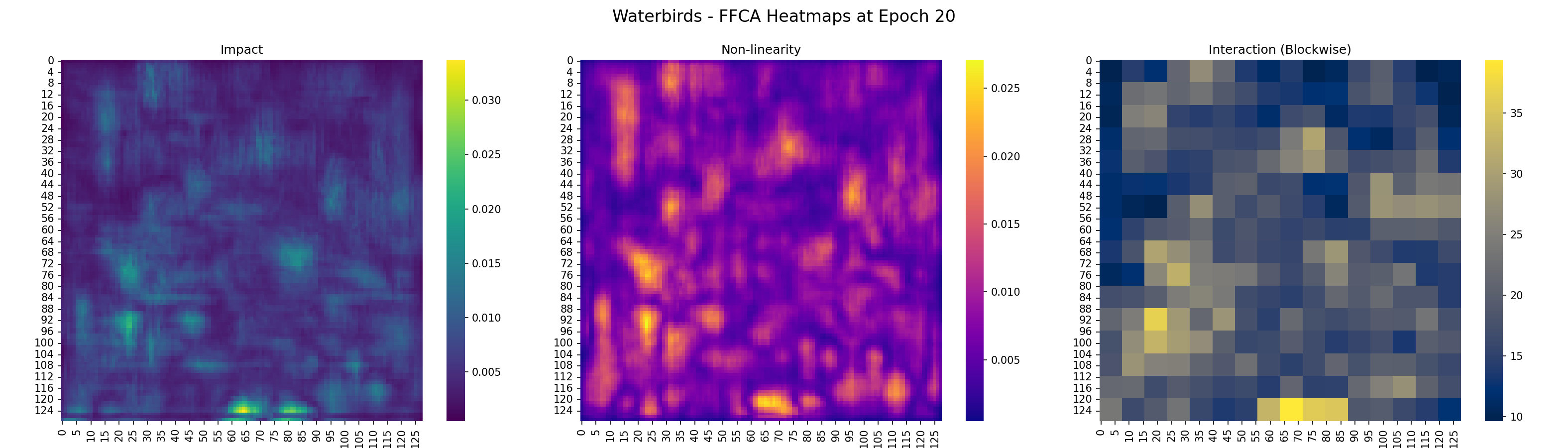}
        \caption{Epoch 20: Attention has drifted to the background.}
    \end{minipage}
    \caption{The evidence of shortcut learning on the Waterbirds dataset. FFCA's aggregate Impact heatmaps show the model's attention shifting from the bird (left) to the background (right) as training progresses. You will need to replace the placeholder images with your saved plots for epoch 5 and 20.}
    \label{fig:waterbirds_drift}
\end{figure}

As shown visually in Figure~\ref{fig:waterbirds_drift} and quantitatively in Table~\ref{tab:waterbirds_drift}, the model initially focuses on the center of the image where the bird is located. However, as training progresses, the model's attention drifts decisively to the background. The FBR collapses from an initial 1.154 to a final 0.356, and our system automatically flags a "Drift Epoch" at epoch 9, where the background reliance becomes dominant. Crucially, this geometric drift directly correlates with the model's failure on the difficult minority cases; minority accuracy stagnates at a low level ($\approx$0.30), proving that the model has failed to learn robust features and has instead converged on the spurious background correlation.

\begin{table}[htbp]
\centering
\caption{Dynamic FFCA metrics reveal the onset of shortcut learning on the Waterbirds dataset. As training progresses, the Foreground-Background Ratio (FBR) collapses and the Center of Mass (CoM) Distance drifts outward, while accuracy on the minority groups stagnates.}
\label{tab:waterbirds_drift}
\begin{tabular}{@{}cccccc@{}}
\toprule
\textbf{Epoch} & \textbf{Val Acc} & \textbf{Minority Acc} & \textbf{CoM Dist} $\uparrow$ & \textbf{FBR} $\downarrow$ & \textbf{Notes} \\ 
\midrule
1 & 0.572 & 0.212 & 4.16 & 1.154 & Initial focus on foreground \\
5 & 0.681 & 0.322 & 3.41 & 0.702 & Performance peaks \\
\textbf{9} & \textbf{0.656} & \textbf{0.299} & \textbf{4.27} & \textbf{0.403} & \textbf{Drift Epoch Detected} \\
15 & 0.643 & 0.299 & 4.93 & 0.388 & Background reliance deepens \\
20 & 0.649 & 0.302 & 4.77 & 0.356 & Final state: shortcut learned \\
\bottomrule
\end{tabular}
\end{table}

\paragraph{Interpreting FFCA Heatmaps for CNNs.}
Although the three FFCA heatmaps (Impact, Non-linearity, Interaction) are visually similar in this case, this is expected in CNNs: regions that strongly influence the output (Impact) typically also exhibit curvature (Non-linearity) once activations are smoothed for analysis, and our interaction estimate is intentionally coarse (blockwise), emphasizing the same large-scale structures. Importantly, the maps encode \emph{different geometric properties}—sensitivity, curvature, and cross-feature synergy—and underpin distinct diagnostics (FBR from Impact; CCI from Non-linearity). We retain all three because agreement across first- and second-order views strengthens the evidence for shortcut learning; conversely, a future \emph{disagreement} between them would localize different failure modes (e.g., linear reliance vs interactive background cues). The apparent similarity is also partly due to per-panel autoscaling of color bars; due to the different scale of each plot's values, we must scale them individually for clarity.

\FloatBarrier
\subsection{Real-world case study: wine classification}
\begin{figure}[htbp]
    \centering
    \includegraphics[width=\columnwidth]{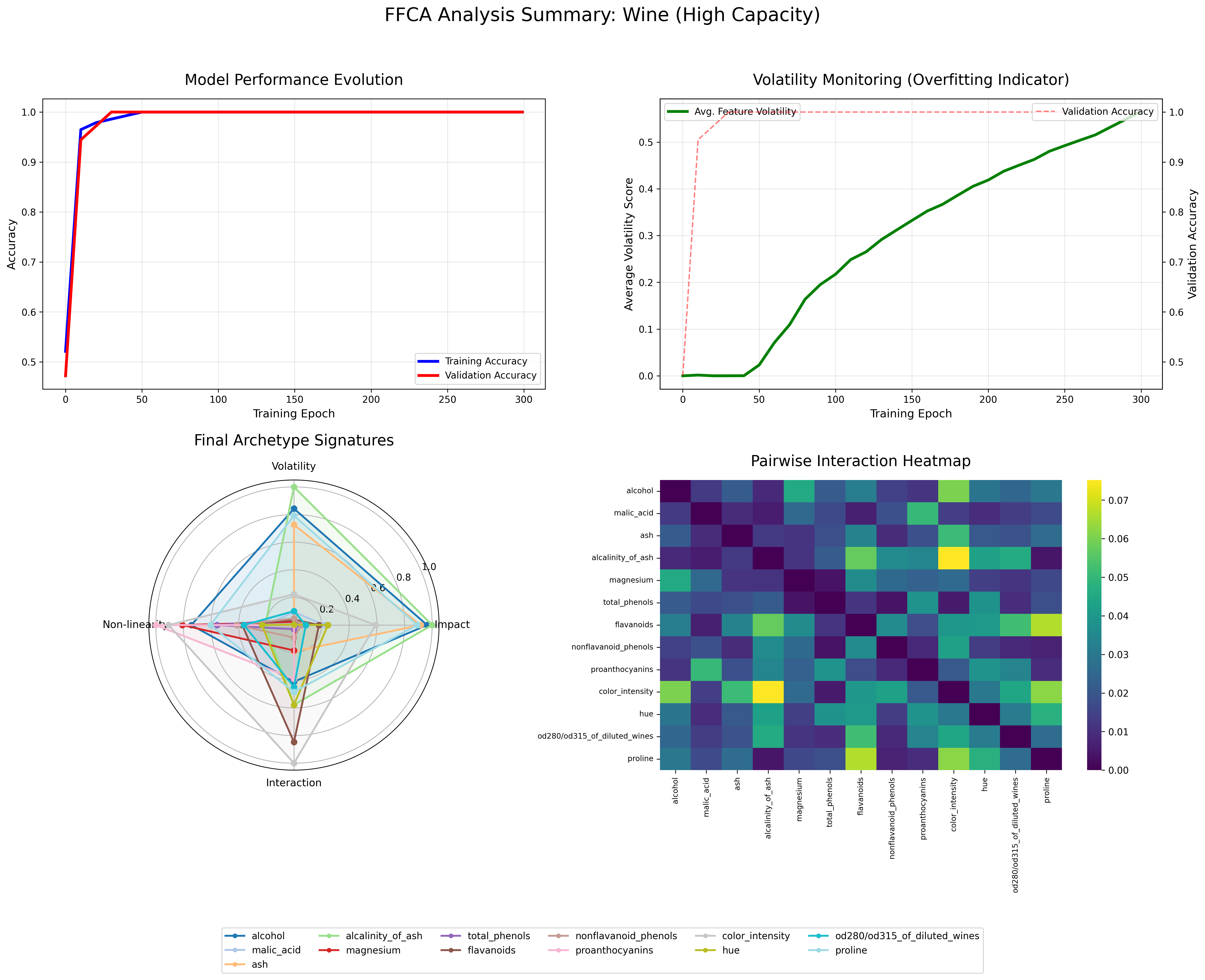}
    \caption{Final analysis for the high-capacity Wine model. The model quickly achieves perfect accuracy (top-left). The final archetypes (bottom-left) reveal the nuanced roles of key chemical components, identifying primary drivers like `alcohol' and `proline' as Volatile Specialists and uncovering the interactive role of `flavanoids'.}
    \label{fig:wine_high_capacity_summary}
\end{figure}

To demonstrate the framework's versatility, we applied Dynamic FFCA to the Wine dataset, a multi-class classification task. Unlike the more complex regression problems, the high-capacity model solves this task with relative ease, reaching 100\% validation accuracy quickly (Figure~\ref{fig:wine_high_capacity_summary}). This ``easy'' case provides a valuable baseline for what efficient learning looks like through the lens of FFCA.

The analysis reveals how the model learns to distinguish between wine cultivars based on chemical composition. The results are particularly insightful from a domain perspective. Several of the most well-known chemical markers, including `alcohol', `proline', and `alcalinity\_of\_ash', are identified not as simple drivers but as Volatile Specialists. This sophisticated finding suggests the model learned that their individual importance is highly contextual and depends on the levels of other constituents.

Furthermore, `flavanoids' is correctly identified as a Hidden Interactor. While it has a modest direct impact, its importance is derived from its synergistic effects with other phenols, a known characteristic in wine chemistry. This case study confirms that even on simpler, solvable tasks, Dynamic FFCA moves beyond basic feature ranking to uncover nuanced, domain-consistent roles and relationships that are critical for a true understanding of the model's logic.

\begin{figure}[htbp]
    \centering
    \includegraphics[width=\columnwidth]{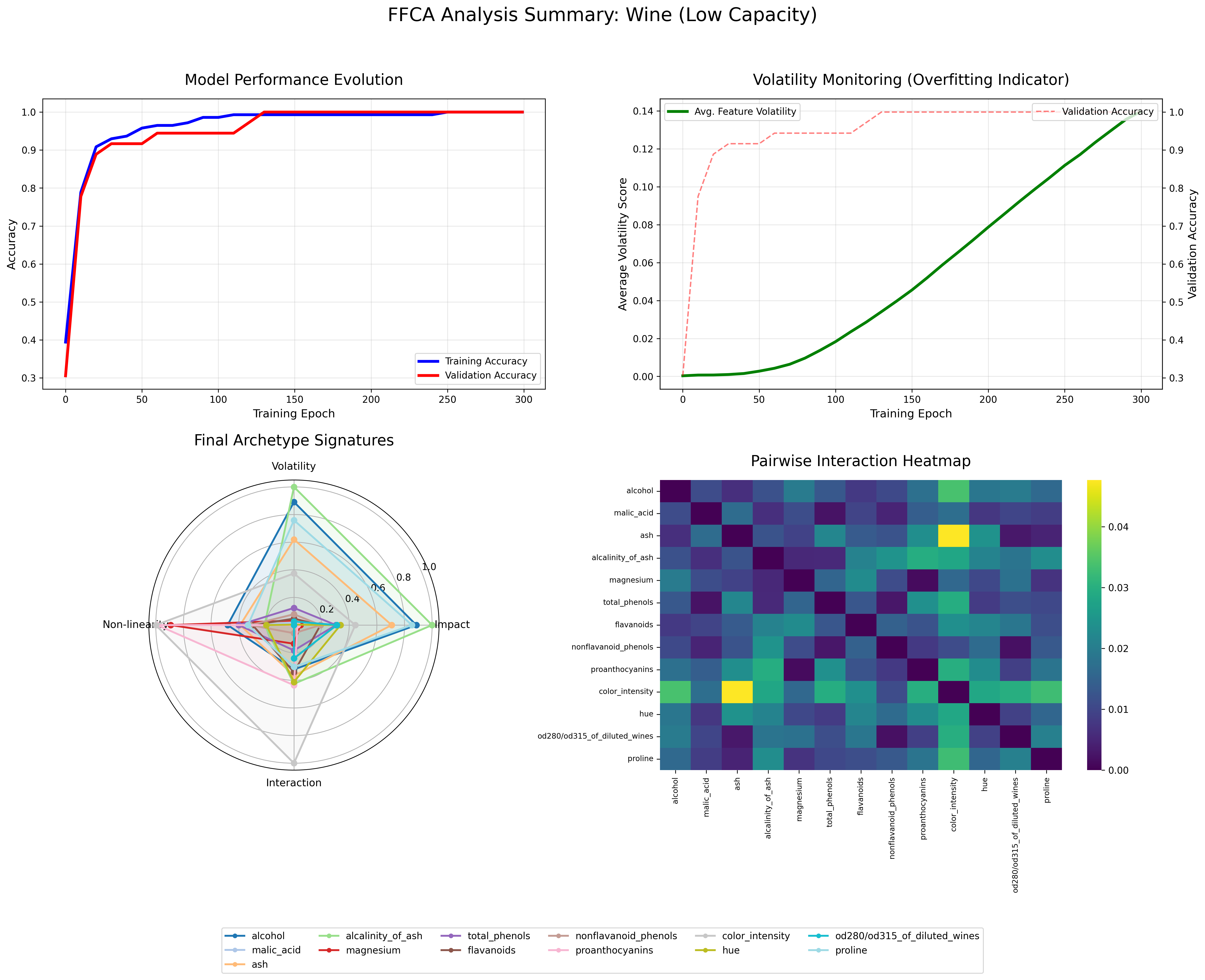}
    \caption{Final analysis for the low-capacity Wine model. Similar to its high-capacity counterpart, this simpler model also achieves perfect accuracy, demonstrating that the dataset's patterns are not overwhelmingly complex.}
    \label{fig:wine_low_capacity_summary}
\end{figure}

The low-capacity model also performs exceptionally well on this dataset, achieving 100\% validation accuracy (Figure~\ref{fig:wine_low_capacity_summary}). This indicates that the underlying relationships in the Wine dataset are strong enough to be captured even by a simpler architecture, providing a useful contrast to the more complex regression tasks.

\subsection{Advanced debugging: detecting spurious correlations and data leakage}
To address the most rigorous critiques of model validation, we conducted two final experiments where we deliberately injected problematic features into the California Housing dataset. These controlled tests were designed to verify if FFCA could produce distinct, identifiable "fingerprints" for spurious correlations and data leakage, two of the most critical and difficult-to-diagnose failure modes in machine learning. The key insight is that while both problems can cause a similar "spike-and-drop" pattern in average feature volatility, the diagnosis becomes clear and unambiguous when interpreting this volatility curve in conjunction with the model's validation performance.

\begin{figure}[htbp]
    \centering
    \includegraphics[width=\columnwidth]{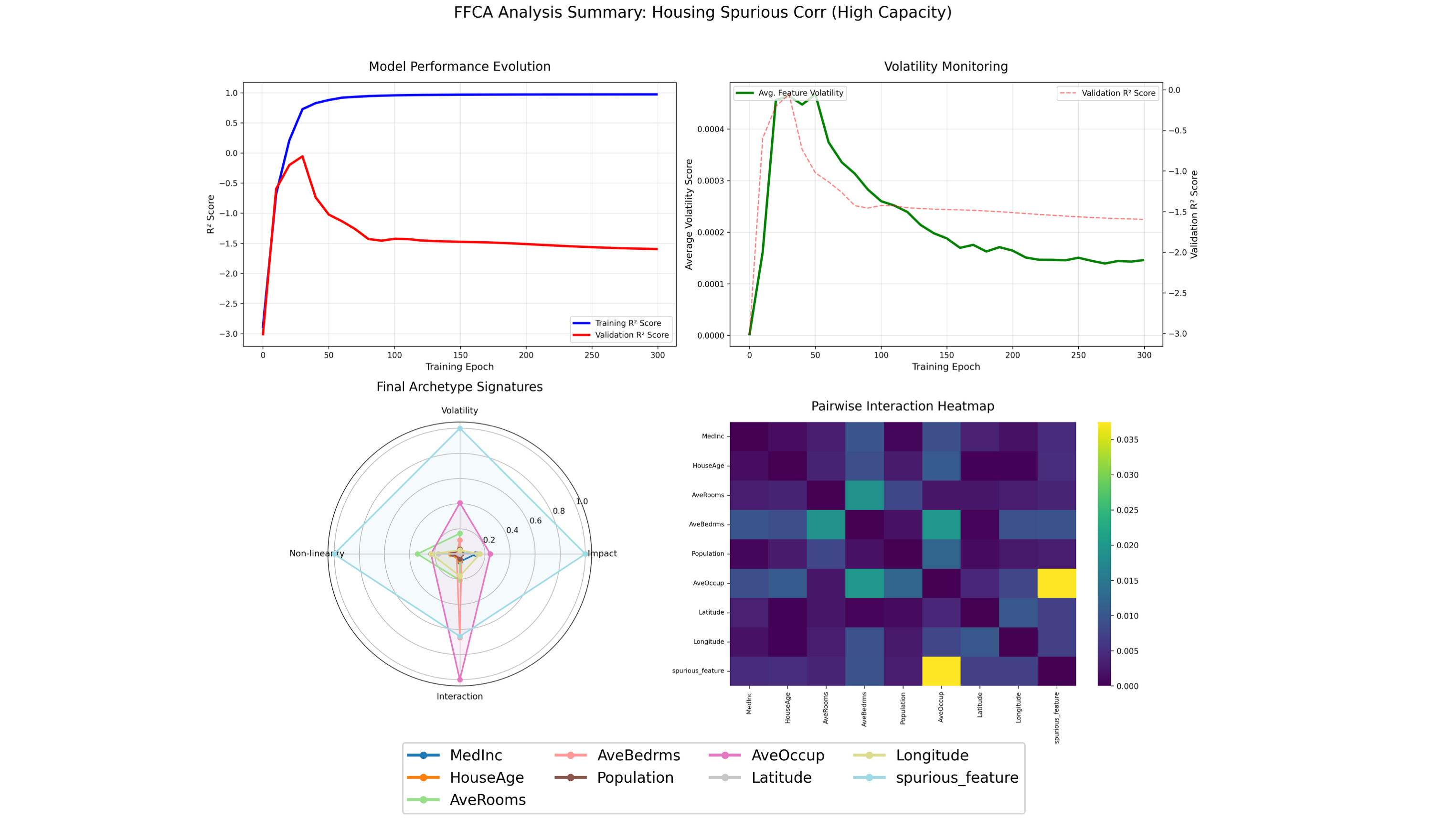}
    \caption{Analysis of the model trained with a spurious feature. \textbf{(Top-Left)} The massive divergence between training (blue) and validation (red) R² scores shows extreme overfitting. \textbf{(Top-Right)} The volatility monitoring plot shows a sharp, early spike as the model learns an unstable pattern, followed by a decline as it converges to an overfitted state. \textbf{(Bottom-Right)} The heatmap shows the model attempting to interact the `spurious\_feature` with `AveOccup'.}
    \label{fig:spurious_corr_summary}
\end{figure}

\paragraph{Detecting spurious correlations.}
In this experiment, we injected a `spurious\_feature` that was engineered to be correlated with the target variable only in the training set, while being random noise in the validation set. The model's performance (Figure~\ref{fig:spurious_corr_summary}) shows a classic case of severe overfitting: it achieves a near-perfect training R² of 0.974 while catastrophically failing on the validation set with an R² of -1.595.

The Volatility Monitoring plot provides the crucial diagnostic. We observe a sharp, early spike in average volatility as the model struggles with the conflicting signals between the training data (where the pattern is strong) and the validation data (where it is noise). The subsequent decline in volatility does not indicate a recovery; rather, it shows the model has "surrendered" to the training data, converging to a stable state of deep overfitting. FFCA's final diagnosis of the `spurious\_feature' as a `Volatile Specialist' is the "fingerprint" of this failure mode. The interaction heatmap reveals another key insight: even as the model is misled, it attempts to relate the `spurious\_feature` to `AveOccup', the most complex of the original features, highlighting how problematic signals can disrupt the learning of genuine relationships.

\begin{figure}[htbp]
    \centering
    \includegraphics[width=\columnwidth]{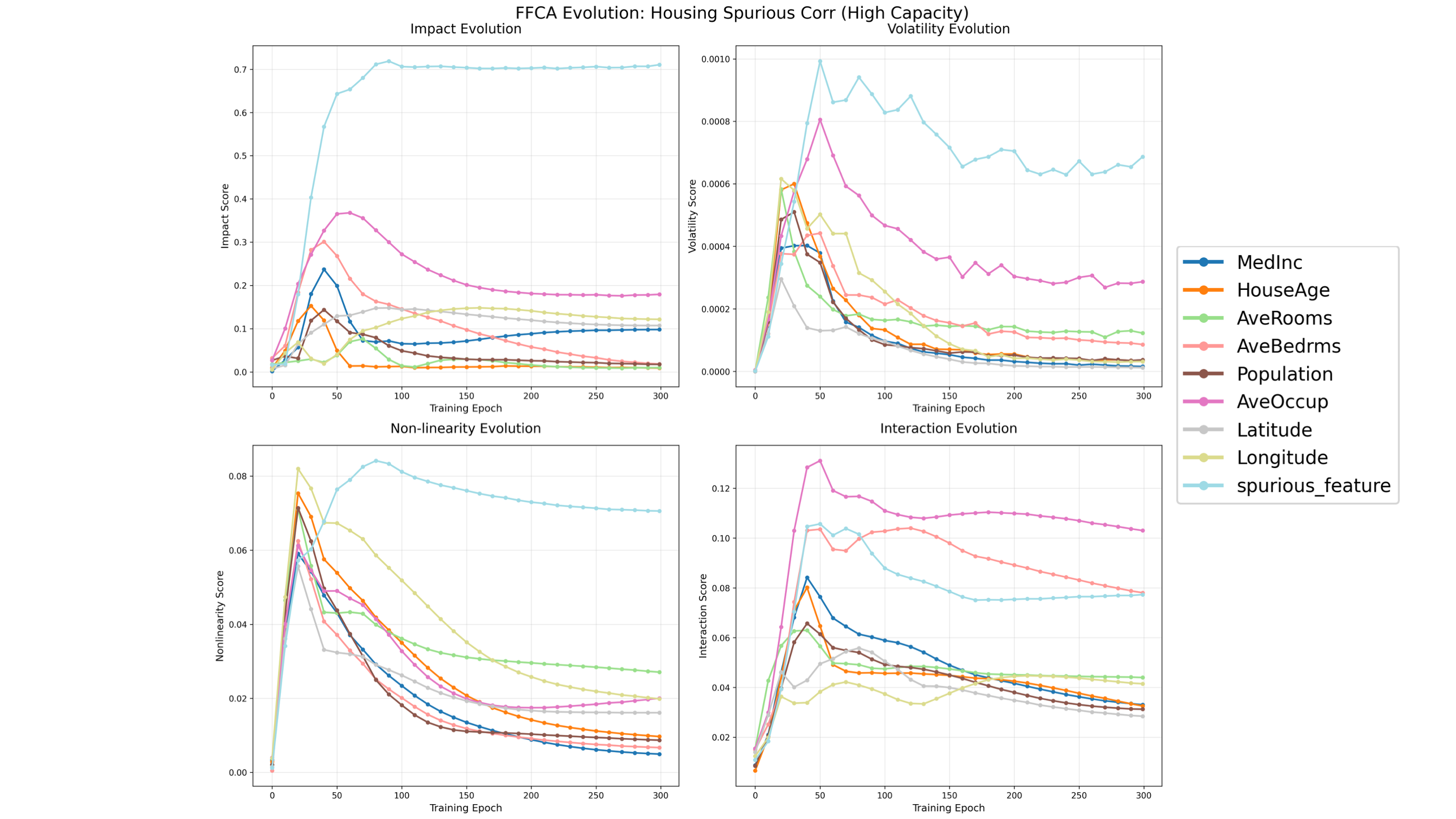}
    \caption{The learning evolution with a spurious feature. The `Impact' of the `spurious\_feature' (purple line, top-left) rises quickly as the model learns the fake pattern. Its `Volatility' (top-right) also climbs as the model struggles with the inconsistent signal between the train and validation sets.}
    \label{fig:spurious_corr_evolution}
\end{figure}

\begin{figure}[htbp]
    \centering
    \includegraphics[width=\columnwidth]{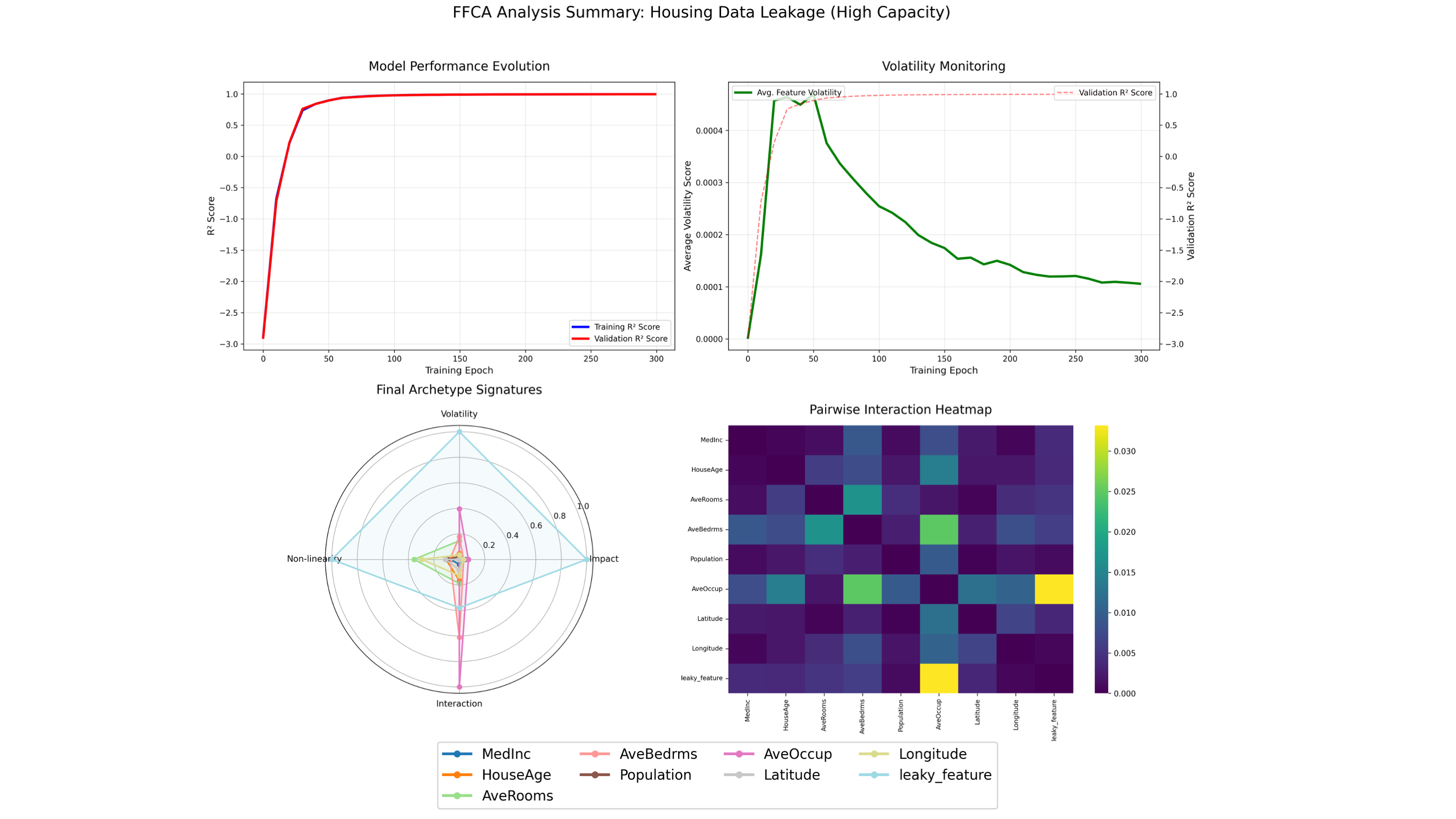}
    \caption{Analysis of the model trained with a leaky feature. (Top-Left)The model achieves unrealistically perfect performance. (Top-Right) The volatility monitoring plot shows a brief spike (the "shock") followed by a steep drop (the "abandonment of learning"), which, combined with the perfect R², confirms a data leak.(Bottom-Left) The radar plot provides the definitive diagnosis: the `leaky\_feature` has a massive Impact score that completely suppresses every other feature.}
    \label{fig:data_leakage_summary}
\end{figure}

\paragraph{Detecting data leakage.}
In the second experiment, we injected a `leaky\_feature' that was a direct, noisy copy of the target variable. The model's performance (Figure~\ref{fig:data_leakage_summary}) immediately signals a problem, achieving an unrealistically perfect R² of 0.996 on both training and validation sets.

\begin{figure}[htbp]
    \centering
    \includegraphics[width=\columnwidth]{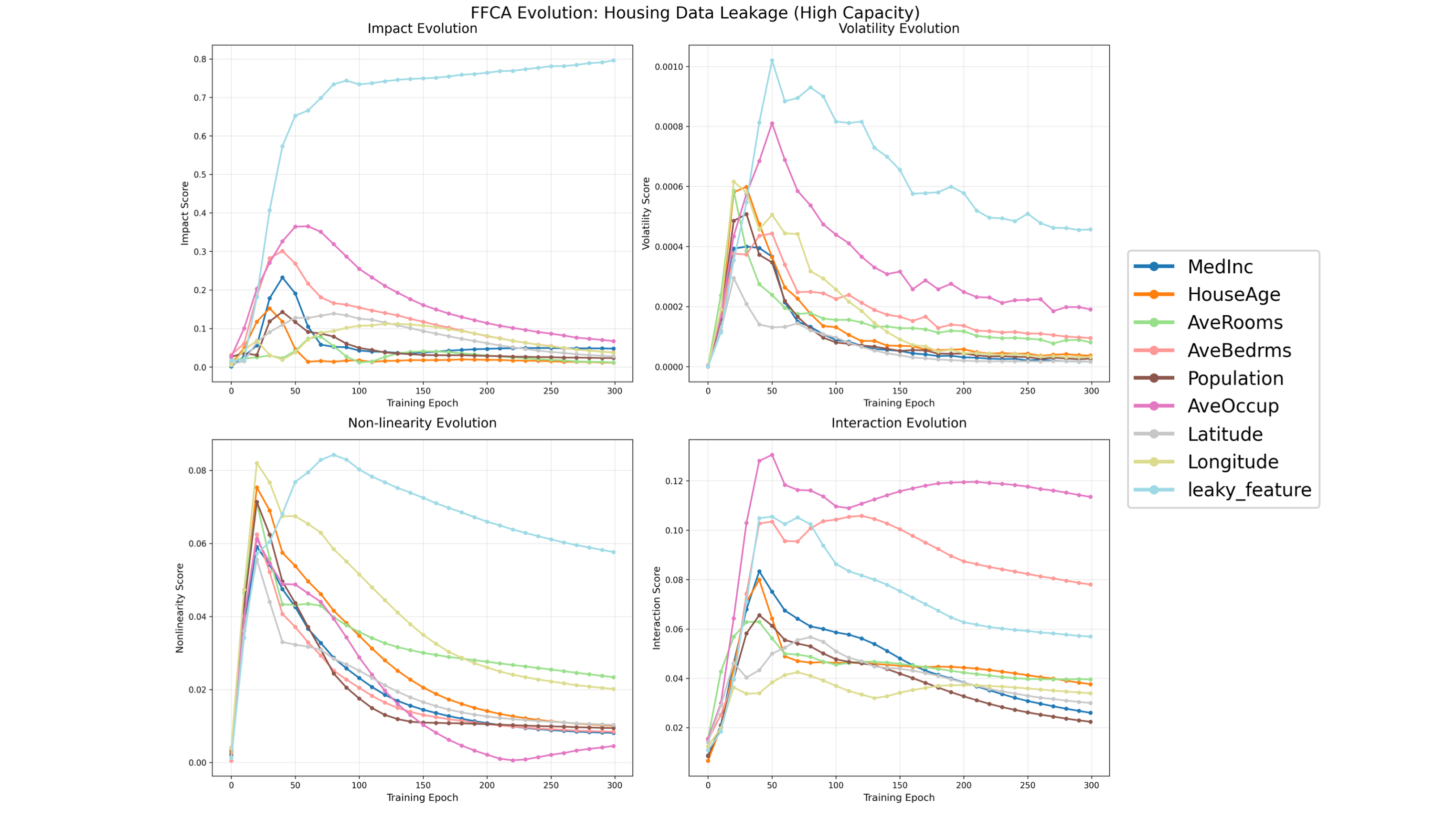}
    \caption{The learning evolution with a leaky feature. The diagnosis is immediate and clear: the `Impact' of the `leaky\_feature' (purple line, top-left) skyrockets from epoch 0, while all other features remain suppressed. This shows the model instantly abandoning the learning process in favor of the leaky signal.}
    \label{fig:data_leakage_evolution}
\end{figure}

The temporal analysis reveals the distinct geometric signature of a leak. The Volatility Monitoring plot again shows a "spike-and-drop" pattern, but its interpretation is different when viewed alongside the perfect validation score. The initial spike represents the geometric "shock" as the model instantly locks onto the overwhelmingly powerful leaky signal. The subsequent steep drop in volatility signifies that the model has completely abandoned the learning process for all other features. It has found the "answer key" and converged to a simple, stable state. This is confirmed in the evolution plot (Figure~\ref{fig:data_leakage_evolution}), where the `Impact' of the `leaky\_feature' explodes from epoch 0 while all other features remain undeveloped. This immediate dominance is the unmistakable FFCA fingerprint of data leakage. The heatmap further shows that the only significant interaction the model considers is between the `leaky\_feature' and `AveOccup', as it tries to reconcile the most complex original feature with the answer key.

\paragraph{Practitioner's guide to diagnosing training pathologies.}
These experiments provide a clear diagnostic workflow for practitioners. When a model exhibits unexpected behavior, the analysis should proceed in three steps. First, examine the \textbf{Model Performance} plot. Does the validation score diverge from the training score (signaling overfitting), or does it reach an unrealistically perfect level (signaling a potential leak)? Second, inspect the \textbf{Volatility Monitoring} plot. A sharp, early spike followed by a decline is a universal sign that the model has locked onto a dominant, simplistic signal. If this "shock and lock" pattern coincides with poor validation performance, it indicates the model has overfitted to a spurious correlation. If it coincides with perfect validation performance, it confirms a data leak. Finally, inspect the final \textbf{FFCA signatures}. In the case of a spurious correlation, the problematic feature will be identified as a `Volatile Specialist'. In the case of a data leak, the `leaky\_feature' will have a massive, dominant `Impact' score that suppresses all other features. This three-step process transforms dynamic FFCA into a powerful tool for not only identifying training problems but also understanding their root cause.

\subsection{Analysis of robustness and scalability}

\paragraph{Robustness.} To test the stability of the FFCA signature, we conducted a model perturbation analysis. We took the trained MLP and added zero-mean Gaussian noise with increasing variance to its weights. The FFCA signatures remained highly stable, with Pearson correlations between the baseline and perturbed signatures exceeding 0.99 even under significant noise. The archetype classifications also proved robust, with over 80\% agreement under realistic noise levels. This demonstrates that FFCA captures fundamental properties of the learned function rather than noisy artifacts of a specific weight configuration.

\paragraph{Scalability.} The default diagonal approximation of FFCA scales linearly with the number of features, $O(d)$, making it highly efficient for computing the Impact, Volatility, and Non-linearity scores. The full approximation, required for the Interaction score, scales as $O(d^2)$. This is more intensive but remains feasible for models with hundreds of features and can be targeted at a pre-selected subset of important features for detailed analysis.


\end{document}